\documentclass[runningheads]{llncs}

\usepackage{eccv}

\usepackage{eccvabbrv}

\usepackage{graphicx}
\usepackage{booktabs}
\usepackage{fontawesome}

\usepackage[accsupp]{axessibility}  

\usepackage{makecell}

\usepackage{hyperref}

\usepackage{orcidlink}

\begin{document}

\title{Deep Learning Meets Satellite Images – An Evaluation on Handcrafted and Learning-based Features for Multi-date Satellite Stereo Images} 

\titlerunning{Deep learning meets satellite images}

\author{Shuang Song\inst{1,2,4}\orcidlink{0000-0002-0037-1499} \and
Luca Morelli\inst{5,6} \orcidlink{0000-0001-7180-2279} \and
Xinyi Wu\inst{1,3}  \orcidlink{0009-0004-4437-7157} \and
Rongjun Qin\inst{1,2,3,4} \textsuperscript{(\faEnvelopeO)} \orcidlink{0000-0002-5896-1379} \and
Hessah Albanwan \inst{7} \and
Fabio Remondino \inst{5} \orcidlink{0000-0001-6097-5342}
}

\authorrunning{S.~Song et al.}

\institute{Geospatial Data Analytics Lab, The Ohio State University, Columbus, USA  \email{\{song.1634, wu.4988, qin.324\}@osu.edu}
\and
Department of Civil, Environmental and Geodetic Engineering, The Ohio State University, Columbus, USA
\and 
Department of Electrical and Computer Engineering, The Ohio State University, Columbus, USA
\and 
Translational Data Analytics Institute, The Ohio State University, Columbus, USA
\and 
3D Optical Metrology (3DOM) Unit, Bruno Kessler Foundation (FBK), Trento, Italy
\email{\{lmorelli,remondino\}@fbk.eu}
\and 
Department of Civil, Environmental and Mechanical Engineering, University of Trento, Italy 
\and
Civil Engineering Department, Kuwait University, Kuwait
\email{hessah.albanwan@ku.edu.kw}\\
}

\maketitle

\begin{abstract}
  A critical step in the digital surface models (DSM) generation is feature matching. Off-track (or multi-date) satellite stereo images, in particular, can challenge the performance of feature matching due to spectral distortions between images, long baseline, and wide intersection angles. Feature matching methods have evolved over the years from handcrafted methods (\eg, SIFT) to learning-based methods (\eg, SuperPoint and SuperGlue). In this paper, we compare the performance of different features, also known as feature extraction and matching methods, applied to satellite imagery. A wide range of stereo pairs ($\sim 500$) covering two separate study sites are used. SIFT, as a widely used classic feature extraction and matching algorithm, is compared with seven deep-learning matching methods: SuperGlue, LightGlue, LoFTR, ASpanFormer, DKM, GIM-LightGlue, and GIM-DKM. Results demonstrate that traditional matching methods are still competitive in this age of deep learning, although for particular scenarios learning-based methods are very promising.
  \keywords{Satellite relative orientation \and Stereo reconstruction \and Multi-temporal image matching}
\end{abstract}

\section{Introduction}
\label{sec:Introduction}
Satellite stereo images are crucial for applications, such as 3D modeling\cite{Chen:2024:0099-1112:371}, mapping \cite{Mezouar:2023:0099-1112:291}, reconstruction \cite{Fu:2023:0099-1112:211, Xu:2022:0099-1112:469}, change detection\cite{Wang:2022:0099-1112:164}, \etc . Their significant advantages are due to their global coverage, low cost per unit area, and frequent revisiting times \cite{bosch_semantic_2019, gui_automated_2021, huang_evaluation_2022}. Current commercial satellites offer images with ground sampling distance (GSD) up to 0.3 meters, potentially producing 1:10,000 topographic maps globally \cite{poli_radiometric_2015, Maune:2023:0099-1112:129}. Most satellite images are collected under less ideal conditions, since they are limited to the orbital track and less flexible satellite steering, making perspective stereo image collection an expensive process. As a result, most of the satellite stereo images are constructed by single images of the same scene collected on separate dates, oftentimes months and years apart, and even from different sensors (satellites). Such images are collected under different sun illuminations, sensor responses, atmospheric conditions, anisotropic surfaces, and seasonal landcover variations, as well as a larger baseline and intersection angle \cite{albanwan_comparative_2022, qin_critical_2019,qin_rpc_2016}. Therefore, satellite stereo pairs from different times/tracks, namely off-track stereo images, face elevated challenges when using traditional (handcrafted) algorithms for feature matching and dense stereo matching \cite{qin_critical_2019}. As a result, the current practice still largely relies on collections that are designated for in-track stereo images, \eg, satellite images taken on the same track and minutes apart, leaving the vast number of satellite images significantly underutilized.

Generally, feature matching methods can be simply categorized as traditional and deep learning-based methods \cite{lindenberger_lightglue_2023, lowe_distinctive_2004, rublee_orb_2011, sarlin_superglue_2020}. Traditional methods are based on handcrafted features (\eg, SIFT \cite{lowe_distinctive_2004}), while deep learning methods (\eg, SuperPoint \cite{detone_superpoint_2018} and SuperGlue \cite{sarlin_superglue_2020}) are trained to handle extreme appearance and viewing angle changing between the stereo pair images. In the last few years, leaning-based approaches have shown consistent progress in image-matching problems and benchmarks \cite{jin_image_2021, remondino_aerial_2022}. Owning to its ability to learn complex features by samples, learning-based methods have shown to be effective in addressing correspondence problems between images with significant differences in scale, illumination, and colorimetry \cite{morelli_deep-image-matching_2024, morelli_photogrammetry_2022}. However, their ability to address the compounded challenges in satellite off-track stereo pairs is just started to be explored \cite{albanwan_comparative_2022,song_evaluating_2024}. In the latter work \cite{song_evaluating_2024}, authors compared the performance of handcrafted and learning-based matching methods on some 40 challenging stereo pairs from ultra-large multi-date satellite image sets by selecting the stereo pairs where the SIFT matcher can find very small number of inliers. 

In this paper, we performed a more thorough study by testing 496 stereo pairs using the 2019 Data Fusion Contest (DFC) \cite{saux_data_2019}. In our evaluation, we consider SIFT as the representative handcrafted method and compare its performance to seven other learning-based matching methods: SuperGlue \cite{sarlin_superglue_2020}, LoFTR \cite{sun_loftr_2021}, ASpanFormer \cite{chen_aspanformer_2022}, LightGlue \cite{lindenberger_lightglue_2023}, DKM \cite{edstedt_dkm_2023}, GIM-LightGlue\cite{shen_gim_2024}, and GIM-DKM \cite{shen_gim_2024}. The performance of the matching algorithms is evaluated by checking the resulting geometric accuracy of the relative orientation, and the accuracy of the generated digital surface model (DSM) against a reference airborne LiDAR dataset.

\section{Related Works}
\label{sec:Related_Works}

Early works on feature extraction and matching in satellite imagery noted the unique challenges of off-track satellite stereo images, while most of them focus on evaluating different dense matching algorithms \cite{albanwan_comparative_2022} or analyzing stereo configurations under varying acquisition conditions \cite{dangelo_high_2014, facciolo_automatic_2017, qin_critical_2019}. For example, \cite{albanwan_comparative_2022} found that end-to-end learning-based dense stereo matching networks can better process off-track stereo images, albeit it may suffer from generalization issues for unseen datasets (\ie, different sensors and resolutions). However, these studies neglected the fact that a feature matcher should be studied in the first place to ensure accurate geo-referencing within a bundle adjustment process. In recent years, new approaches based on convolutional neural networks (CNNs) have been proposed to overcome the limitations of traditional handcrafted local features, such as SIFT \cite{lowe_distinctive_2004} and ORB \cite{rublee_orb_2011}. Conventional methods exhibit suboptimal performance when matching images characterized by substantial variations in illumination conditions and/or viewing angles. Typically, these CNNs are trained via self-supervised techniques, utilizing multi-temporal datasets derived from diverse sensors and including a broad spectrum of objects and environments \cite{detone_superpoint_2018}. Detection and description have been trained separately, \eg Key.Net \cite{barroso-laguna_keynet_2019} and HardNet \cite{mishchuk_working_2017}, or jointly, as in SuperPoint \cite{detone_superpoint_2018}. Concurrently, there is a growing trend towards employing learning-based methods, such as SuperGlue \cite{sarlin_superglue_2020} and LightGlue \cite{lindenberger_lightglue_2023}, among others. For an overview of deep-learning local features and accuracy evaluation, see \cite{jin_image_2021,morelli_photogrammetry_2022} for more details.

Recently, differing from key point and feature descriptor-based matching methods, detector-free matching processes a pair of images and output correspondences in one shot \cite{chen_aspanformer_2022}. LoFTR \cite{sun_loftr_2021} first skips keypoint detection, employing transformers for global matching to succeed in low-texture areas. ASpanFormer \cite{chen_aspanformer_2022} introduces an adaptive span transformer and was pre-trained to address both low-texture and large perspective changes. DKM \cite{edstedt_dkm_2023} introduces a dense kernelized feature matching approach that significantly improves two-view geometry estimation. These detector-free methods offer dense and evenly distributed correspondences compared to key point techniques, making them particularly suitable for satellite relative orientation tasks.

As mentioned before, classical photogrammetric images are collected at an ideal condition, \ie, with minimal illumination problems and perspective distortions. Therefore, the adoption of learning-based approaches offers fewer advantages than in challenging cases, and, sometimes even results in reduced accuracy, as reported by \cite{remondino_evaluating_2021}. The advantage of the learning-based method, instead, is evident in challenging multi-temporal datasets \cite{maiwald_fully_2021, morelli_photogrammetry_2022} or under different viewing angles \cite{ioli_replicable_2023}. It is noteworthy that these approaches have inherent constraints, including the ability to execute predictions solely on images of limited dimensions determined by GPU capabilities, as well as limitations in rotation and scale invariance, as observed in \cite{marelli_enrich_2023}.

\section{Methodology}
\label{sec:Methodology}

\subsection{The Proposed Processing and Evaluation Framework}
\label{sec:Methodology3.1}
\begin{figure}[hbt!]
  \centering
  \includegraphics[height=4.4cm]{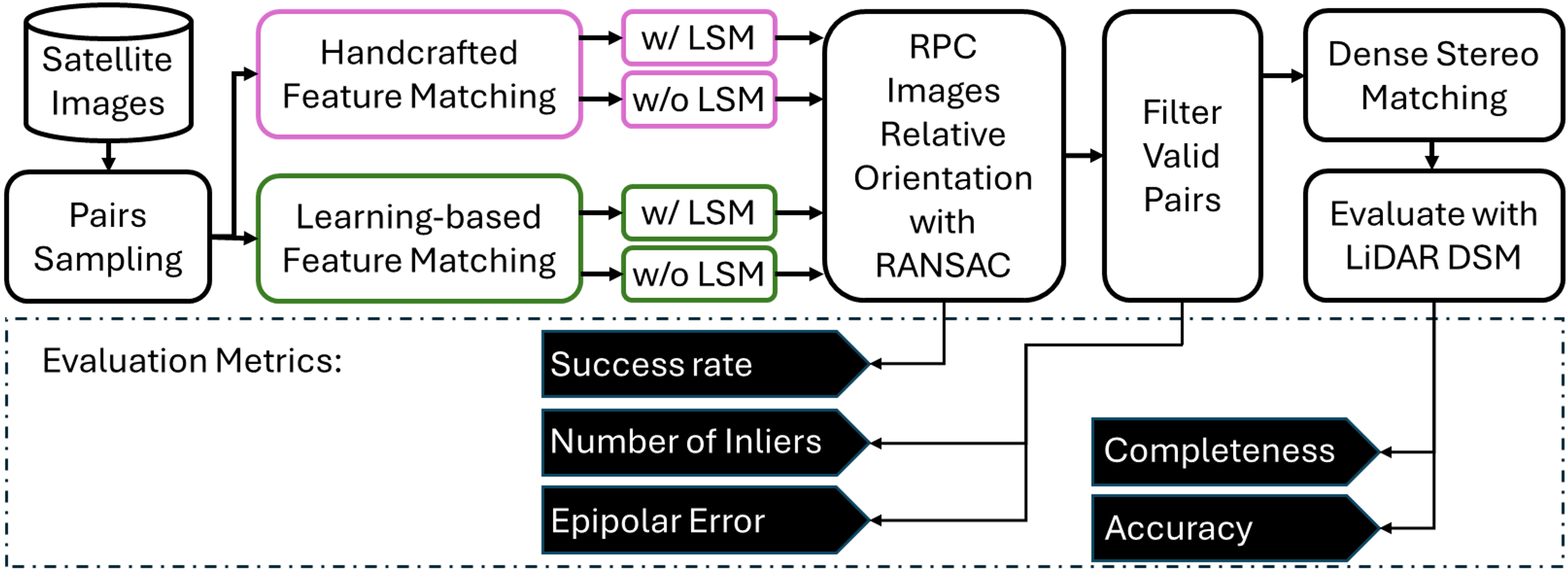}
  \caption{The evaluation workflow}
  \label{fig:The_evaluation_workflow}
\end{figure}

The processing and evaluation framework, shown in \cref{fig:The_evaluation_workflow}, aims to assess the performance of classic handcrafted and learning-based feature matching methods. Firstly, the satellite off-track stereo pairs are selected with proper convergence angle and a challenging appearance difference(see \cref{sec:Methodology3.2}), from which correspondences have been identified by both traditional (\ie, SIFT) and learning-based features (see \cref{sec:Methodology3.3}). Considering that the localization accuracy of different methods varies, we refine these identified matches using Least Squares Matching (LSM) \cite{bellavia_progressive_2024, bethmann_least-squares_2010, gruen_adaptive_1985}. Using these point correspondences, a RPC-based (Rational Polynomial Coefficients) relative orientation/bias compensation is performed using the RSP (RPC stereo processor) software \cite{qin_rpc_2016} RSP incorporates RANSAC and adjusts the RPC coefficients for the image pairs. Our evaluation is based on the success rate of relative orientation, the number of correctly matched points (inliers), and the epipolar error (y-parallax in the epipolar space) (see \cref{sec:Methodology3.4}). We set a threshold in epipolar error to filter out successfully corrected RPC coefficients which are good for dense stereo reconstruction. To ensure fairness in comparison, statistics are all based on those successful stereo pairs.

In addition, the accuracy of the successively computed DSM is also assessed. After completing the relative orientation step, dense stereo matching is performed to create a DSM using the RSP software \cite{qin_rpc_2016}, which implements a typical SGM (Semi-Global Matching) algorithm \cite{hirschmuller_stereo_2008}. The reconstructed DSM is compared to a 3D ground truth DSM, created from an airborne LiDAR sensor \cite{saux_data_2019}, using completeness and accuracy (see \cref{sec:Methodology3.4}).

\subsection{Satellite Off-track Stereo Pairs - Data Preparation}
\label{sec:Methodology3.2}
\begin{figure}[hbt!]
  \centering
  \begin{subfigure}{0.45\linewidth}
    \centering
    \includegraphics[width=\linewidth]{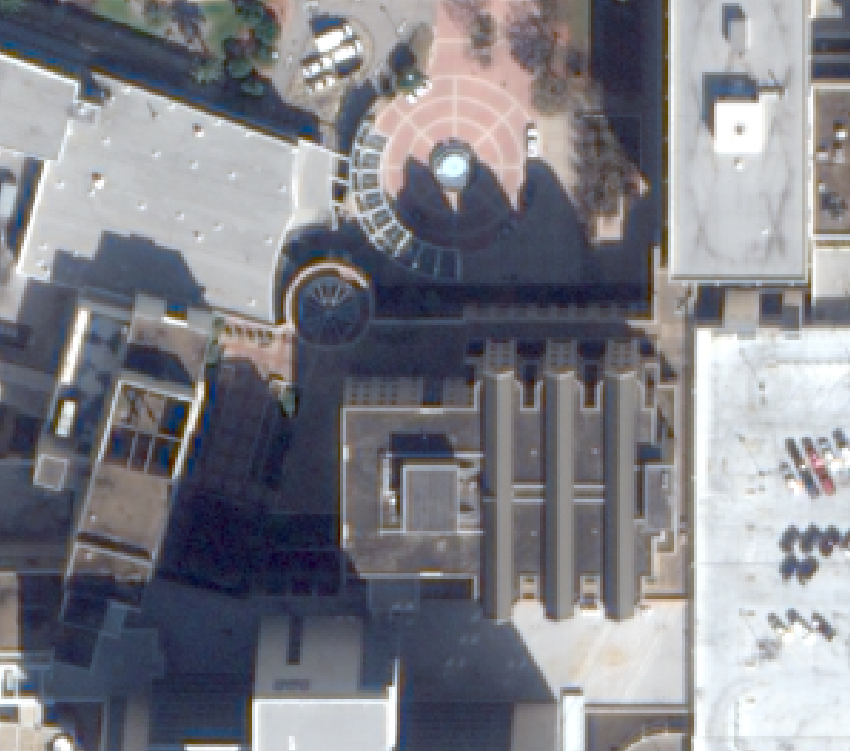}
    \caption{Sun Azimuth 150.2°\newline Elevation 42.2°}
    \label{fig:An_example_of_illumination_difference_JAX_FL-a}
  \end{subfigure}
  \hfill
  \begin{subfigure}{0.45\linewidth}
    \centering
    \includegraphics[width=\linewidth]{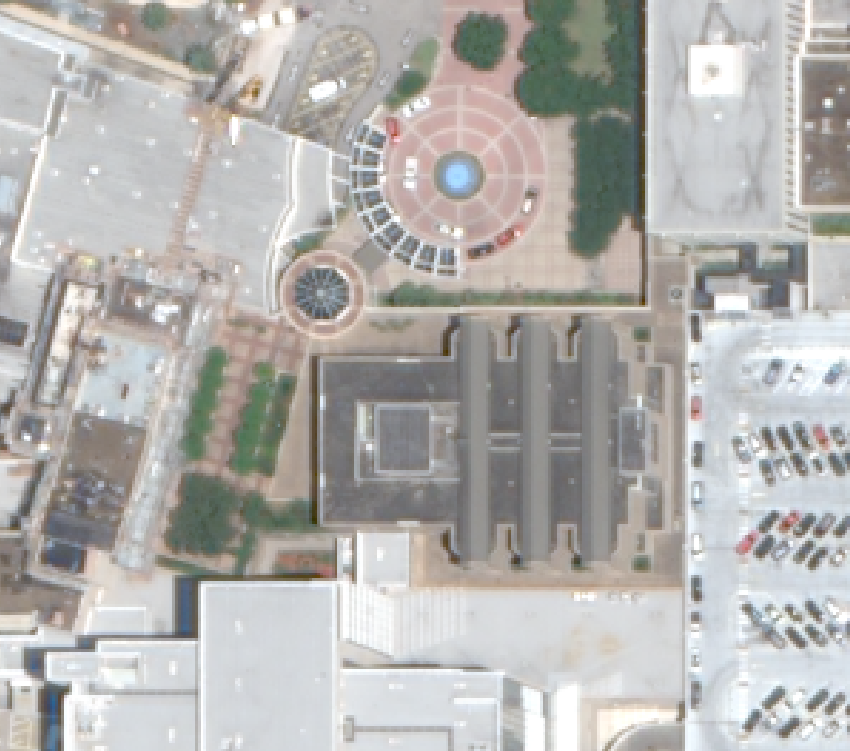}
    \caption{Sun Azimuth 108.0°\newline Elevation 72.1°}
    \label{fig:An_example_of_illumination_difference_JAX_FL-b}
  \end{subfigure}
  \caption{An example of illumination difference (JAX, FL)}
  \label{fig:An_example_of_illumination_difference_JAX_FL}
\end{figure}

\begin{figure}[hbt!]
  \centering
  \begin{subfigure}{0.45\linewidth}
    \centering
    \includegraphics[width=\linewidth]{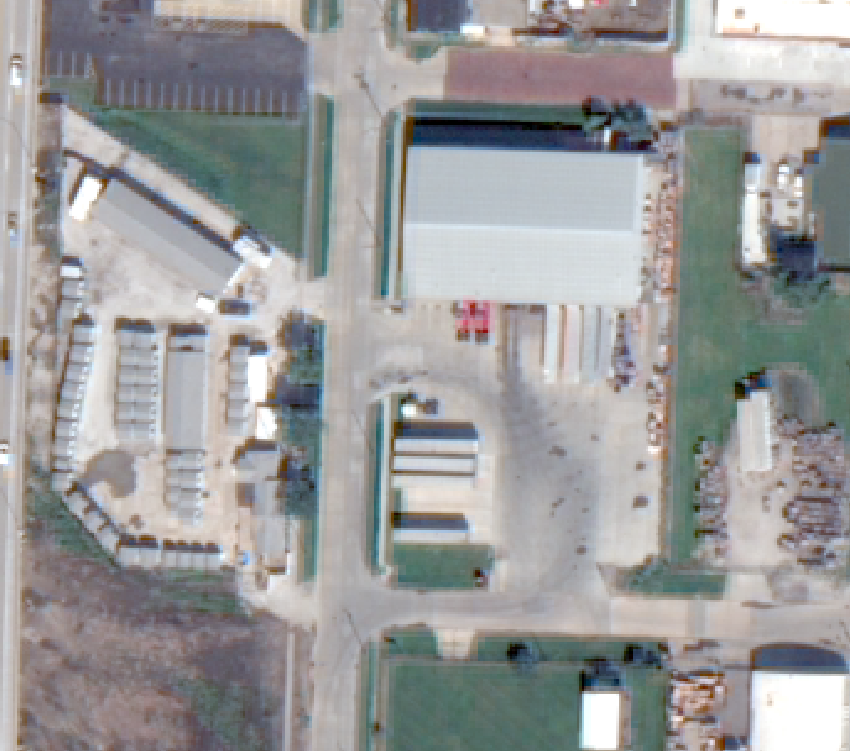}
    \caption{Sept 2014}
    \label{fig:An example of seasonal difference (OMA, NE)-a}
  \end{subfigure}
  \hfill
  \begin{subfigure}{0.45\linewidth}
    \centering
    \includegraphics[width=\linewidth]{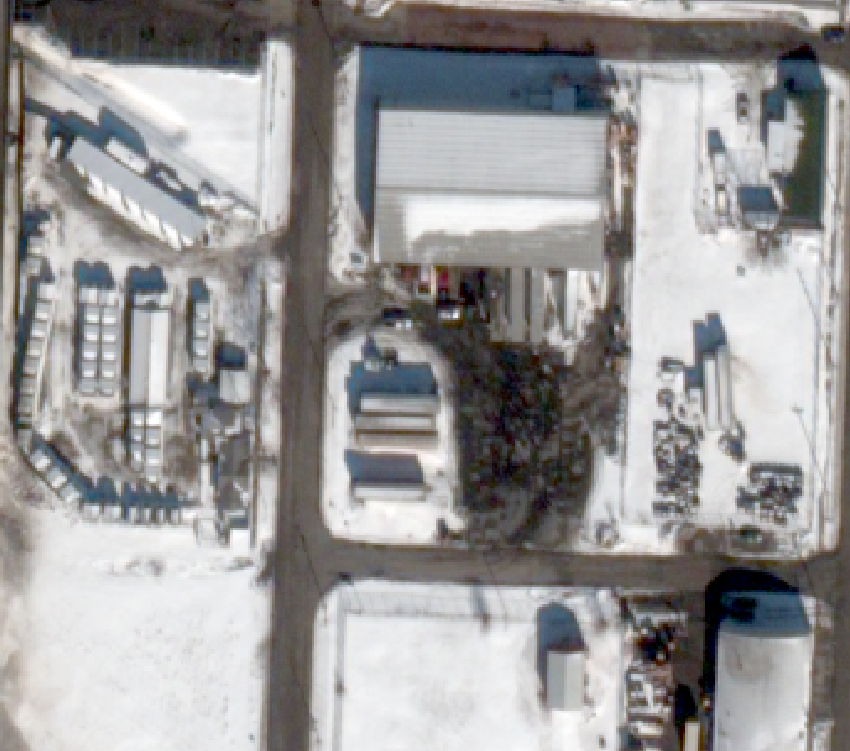}
    \caption{Feb 2015}
    \label{fig:An example of seasonal difference (OMA, NE)-b}
  \end{subfigure}
  \caption{An example of seasonal difference (OMA, NE)}
  \label{fig:An example of seasonal difference_OMA_NE}
\end{figure}

Classic feature matching with handcrafted approaches, such as SIFT, has been widely used in aerial / satellite photogrammetry because of their robustness and efficiency \cite{ling_unified_2021}. However, as mentioned earlier, it falls short in cases where drastic illumination, scale, and/or view differences are observed. Our evaluation focuses on these challenging cases where images show significant appearance differences. To derive 3D geometry, we select stereo pairs with specific intersection angles in the range of 5° to 35° \cite{albanwan_comparative_2022, qin_critical_2019, qin_rpc_2016}. These selected stereo pairs are ranked based on their seasonal and sun illumination differences, \ie, sun angle difference and month-of-year difference using attributes from metadata, respectively. An example where illumination change leads to a huge difference in appearance is shown in \cref{fig:An_example_of_illumination_difference_JAX_FL} whereas seasonal differences are shown in \cref{fig:An example of seasonal difference_OMA_NE}. 
The month-of-year difference is computed with \cref{eq:month-of-year_difference}, where $month_i$ refers to the month-of-year of two paired images.

\begin{equation}
  \min(|month_1 - month_2|, 12-|month_1 - month_2|).
  \label{eq:month-of-year_difference}
\end{equation}


After applying the intersection angle criteria, we randomly select $K$ pairs from the pair pool of each tile, where $K=5$ in our evaluation. 

\subsection{Pair Matching with Handcrafted and Learning-based Features and Matchers}
\label{sec:Methodology3.3}
As a popular descriptor in academia and industry for the last few decades, SIFT is selected as the representative for the handcrafted method. Learning-based methods, start from the milestone SuperPoint/SuperGlue which was introduced in 2020.  \cref{tab:Image Matching methods for evaluation} reports the employed methods. For matching SIFT features, the classic nearest neighbor approach is used with a ratio threshold equal to 0.95 instead of 0.80-0.85. Indeed, preliminary tests have shown that on these datasets affected by extreme seasonal and illumination changes, a too low ratio threshold is too restrictive in discarding ambiguous matches. With a larger threshold, more matches are retained, leaving the elimination of possible outliers to the test with epipolar geometry. SuperPoint and X-Glue follow the sparse key point detection and feature description stages. SuperGlue and LightGlue are two matching methods based on features extracted by SuperPoint. These algorithms are available in the DIM (Deep-Image-Matching) library \cite{morelli_deep-image-matching_2024, morelli_photogrammetry_2022}.

Detector-free matching methods use a different protocol, which does not require a key point detector. Those methods yield semi-dense matchings instead of sparse keypoints. We include LoFTR, ASpanFormer, and DKM  in the detector-free category. GIM \cite{shen_gim_2024} is a self-training method for image matching methods, which provides weights of DKM and LightGlue trained using internet videos using self-training schema. Our evaluation includes the LightGlue and DKM networks trained with GIM method and denoted as GIM-LightGlue, and GIM-DKM respectively.

\begin{table}[hbt!]
  \caption{Image Matching methods for evaluation}
  \label{tab:Image Matching methods for evaluation}
  \centering
  \begin{tabular}{@{\hskip 0.0in}l@{\hskip 0.8in}l@{\hskip 0.8in}l@{\hskip 0.0in}}
    \toprule
    \bf Method  & \bf Method Type & \bf Category \\
    \toprule
    SIFT (1999) \cite{lowe_distinctive_2004}          & Handcrafted & Detector-based \\
    \midrule
    SuperGlue (2020) \cite{sarlin_superglue_2020}     & Learning-based & Detector-based \\
    LoFTR (2021) \cite{sun_loftr_2021}                & Learning-based & Detector-free \\
    ASpanFormer (2022) \cite{chen_aspanformer_2022}   & Learning-based & Detector-free \\
    LightGlue (2023)\cite{lindenberger_lightglue_2023}& Learning-based & Detector-based \\
    DKM (2023) \cite{edstedt_dkm_2023}                & Learning-based & Detector-free \\
    GIM-LightGlue (2024) \cite{shen_gim_2024}         & Learning-based & Detector-based \\
    GIM-DKM (2024) \cite{shen_gim_2024}               & Learning-based & Detector-free \\
    \bottomrule
  \end{tabular}
\end{table}

\subsection{Evaluation Metrics}
\label{sec:Methodology3.4}
As described in \cref{sec:Methodology3.1}, the evaluation metrics are twofold: \emph{(1)} statistics following RPC-based relative orientation and \emph{(2)} a comparison of dense reconstruction to the ground truth DSM.

Our first metric is based on the statistics of relative orientation, including the success rate, the inlier ratio and the epipolar error of the inliers. Instead of adjusting full RPC parameters (80 coefficients in total), we employed 1\textsuperscript{st}-order bias correction similar to a previous work \cite{qin_rpc_2016}. The inlier ratio indicates the number of inliers after RANSAC the initial number of matchings and assesses the effectiveness and precision of the feature matching process. A larger number of inliers increases our confidence in the relative orientation results, as it suggests a smaller number of erroneous matches. The epipolar errors (y-parallax) of inliers are calculated, which means for each matched point, the distance in pixels between a matched point and its corresponding epipolar line. We use the root mean squared epipolar error of all valid matches as a metric, with a smaller error indicating a better matching quality. This metric has been particularly useful in evaluating matching quality when the number of inliers is too low to warrant a reliable relative orientation, potentially impacting the accuracy of the subsequent dense image matching and DSM generation. We use an empirical threshold $T=5 px$ for root mean squared epipolar error in our evaluation to ensure the quality of dense stereo matching. Any pair that has a greater value than the threshold is marked as relative orientation failure and will not proceed in comparison and further processes (\ie, dense stereo matching). Therefore, to ensure a fair comparison, we excluded failure pairs in statistics on both the relative orientation stage and dense reconstruction stage.

For image pairs where both classic and learning-based methods provide enough matching points for reliable orientation, we assess the RPCs' quality by creating a DSM through dense stereo matching and comparing it to the actual ground truth DSM. In this scenario the metric is composed by the completeness and the accuracy of the resulting DSM. The completeness of the DSM is defined as the percentage of the ground truth DSM's area that the derived DSM covers. Completeness values range from 0\% to 100\%, with values closer to 100\% indicating superior dense reconstruction. The accuracy of the DSM is the RMSE (Root Mean Square Error) between the derived DSM and the ground truth DSM. To eliminate the possible systematic error due to the misalignment of generated and ground truth DSMs, we apply least squares surface matching to DSMs \cite{qin_rpc_2016}. Then, the RMSE of pixel-wise distances is computed in co-registered DSMs, excluding pixels classified as NaN (Not a Number) from both the generated and ground truth DSM.

\section{Experiments and Evaluation}
\label{sec:Experiments}
\subsection{Datasets}
\label{sec:Experiments4.1}

\begin{figure}[hbt!]
  \centering
  \begin{subfigure}{0.9\linewidth}
    \centering
    \includegraphics[height=1.2cm]{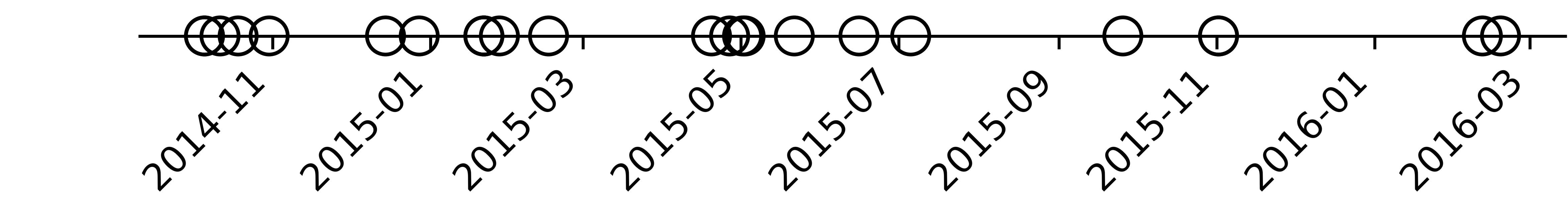}
    \caption{JAX, FL (24 images)}
    \label{fig:Image collection time of DFC2019 dataset-a}
  \end{subfigure}
  \hfill
  \begin{subfigure}{0.9\linewidth}
    \centering
    \includegraphics[height=1.2cm]{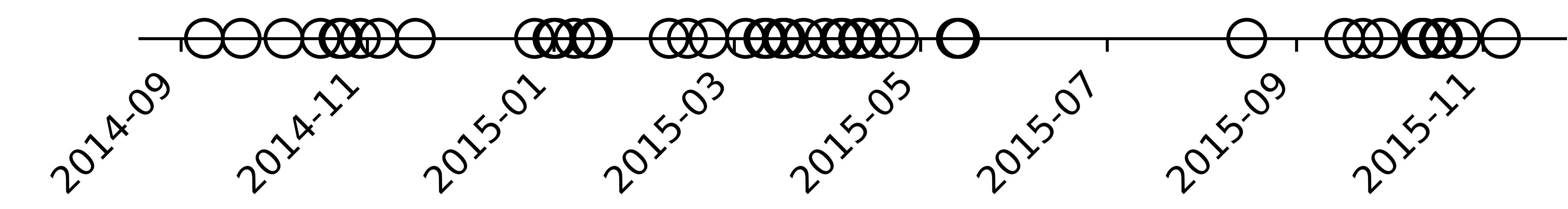}
    \caption{OMA, NE (43 images)}
    \label{fig:Image collection time of DFC2019 dataset-b}
  \end{subfigure}
  \caption{Image collection time within the DFC2019 dataset}
  \label{fig:Image collection time of DFC2019 dataset}
\end{figure}

\begin{figure}[hbt!]
  \centering
  \begin{subfigure}{0.45\linewidth}
    \centering
    \includegraphics[height=4.5cm]{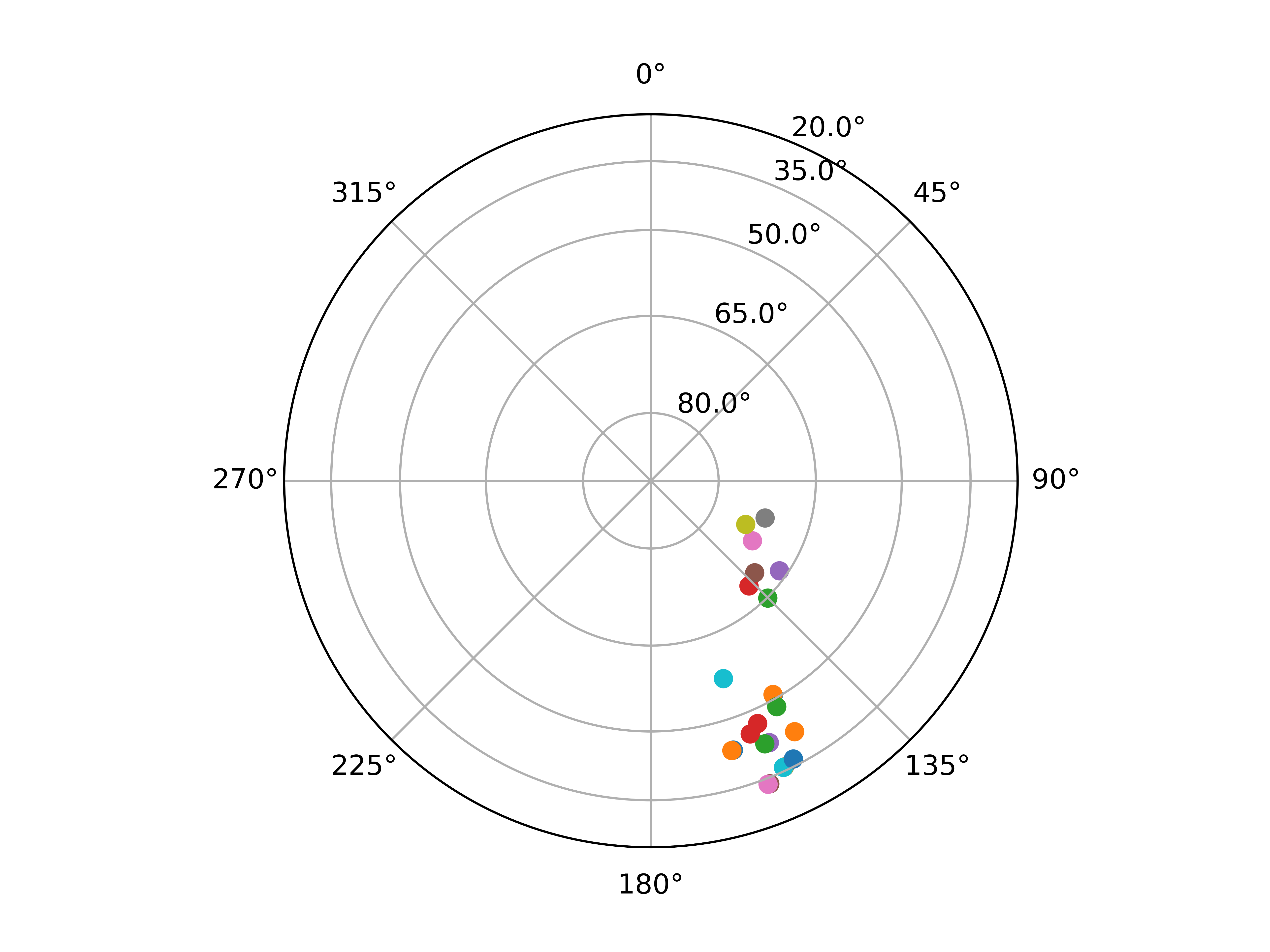}
    \caption{Sun direction (JAX, FL)}
    \label{fig:Imaging properties of DFC2019 dataset-a}
  \end{subfigure}
  \hfill
  \begin{subfigure}{0.45\linewidth}
    \centering
    \includegraphics[height=4.5cm]{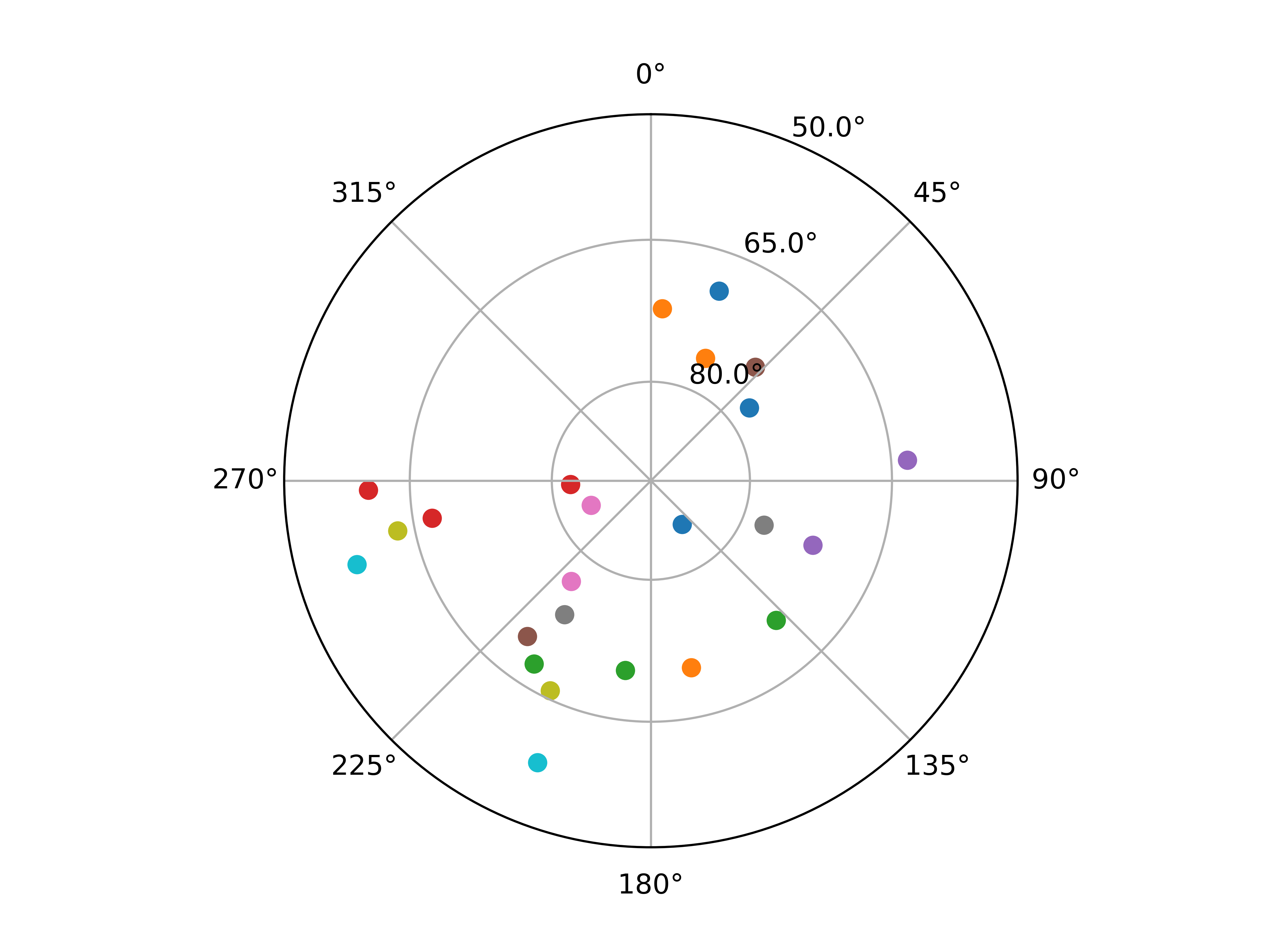}
    \caption{Viewing direction (JAX, FL)}
    \label{fig:Imaging properties of DFC2019 dataset-b}
  \end{subfigure}
  \vfill
  \begin{subfigure}{0.45\linewidth}
    \centering
    \includegraphics[height=4.5cm]{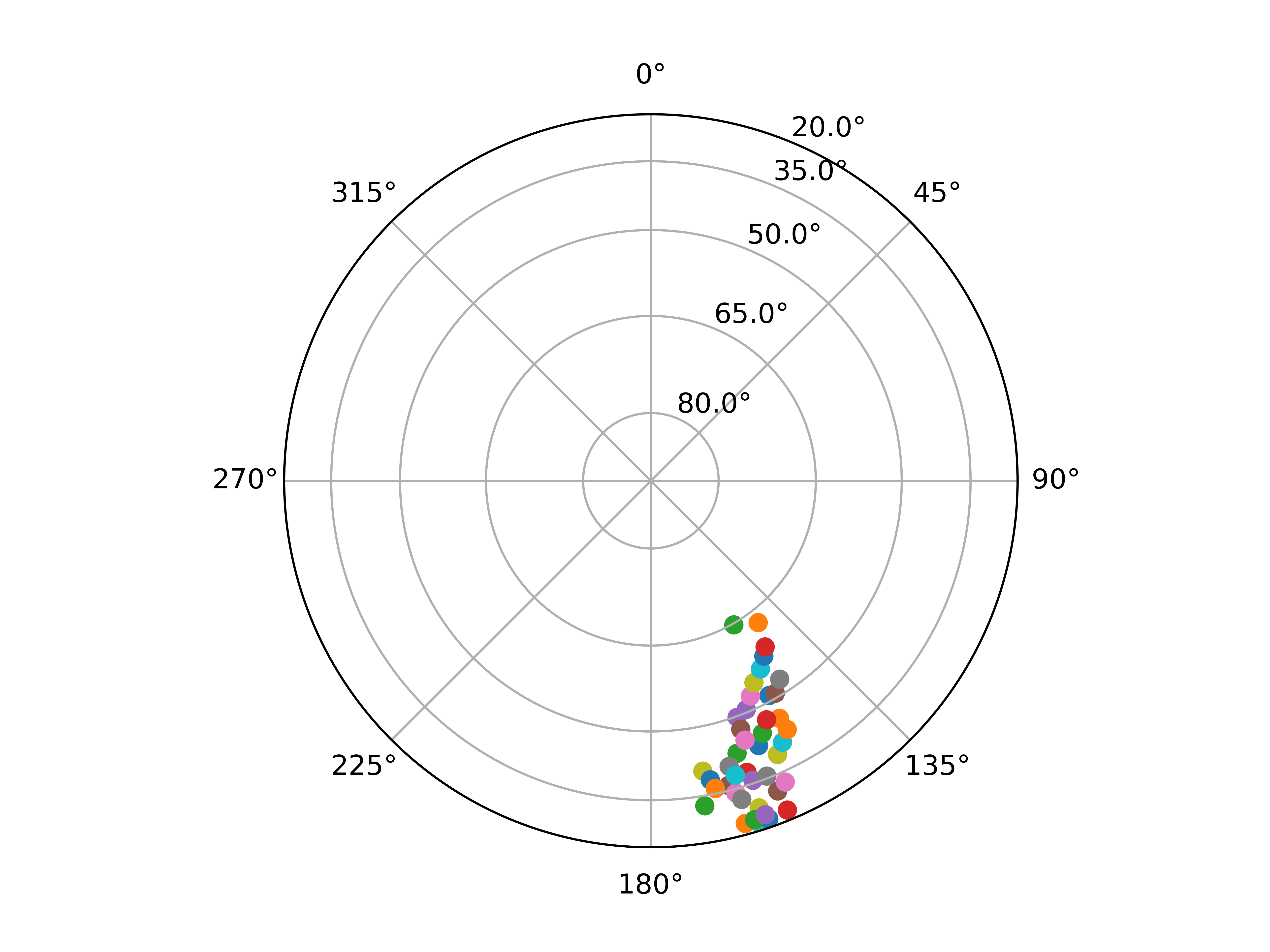}
    \caption{Sun direction (OMA, NE)}
    \label{fig:Imaging properties of DFC2019 dataset-c}
  \end{subfigure}
  \hfill
  \begin{subfigure}{0.45\linewidth}
    \centering
    \includegraphics[height=4.5cm]{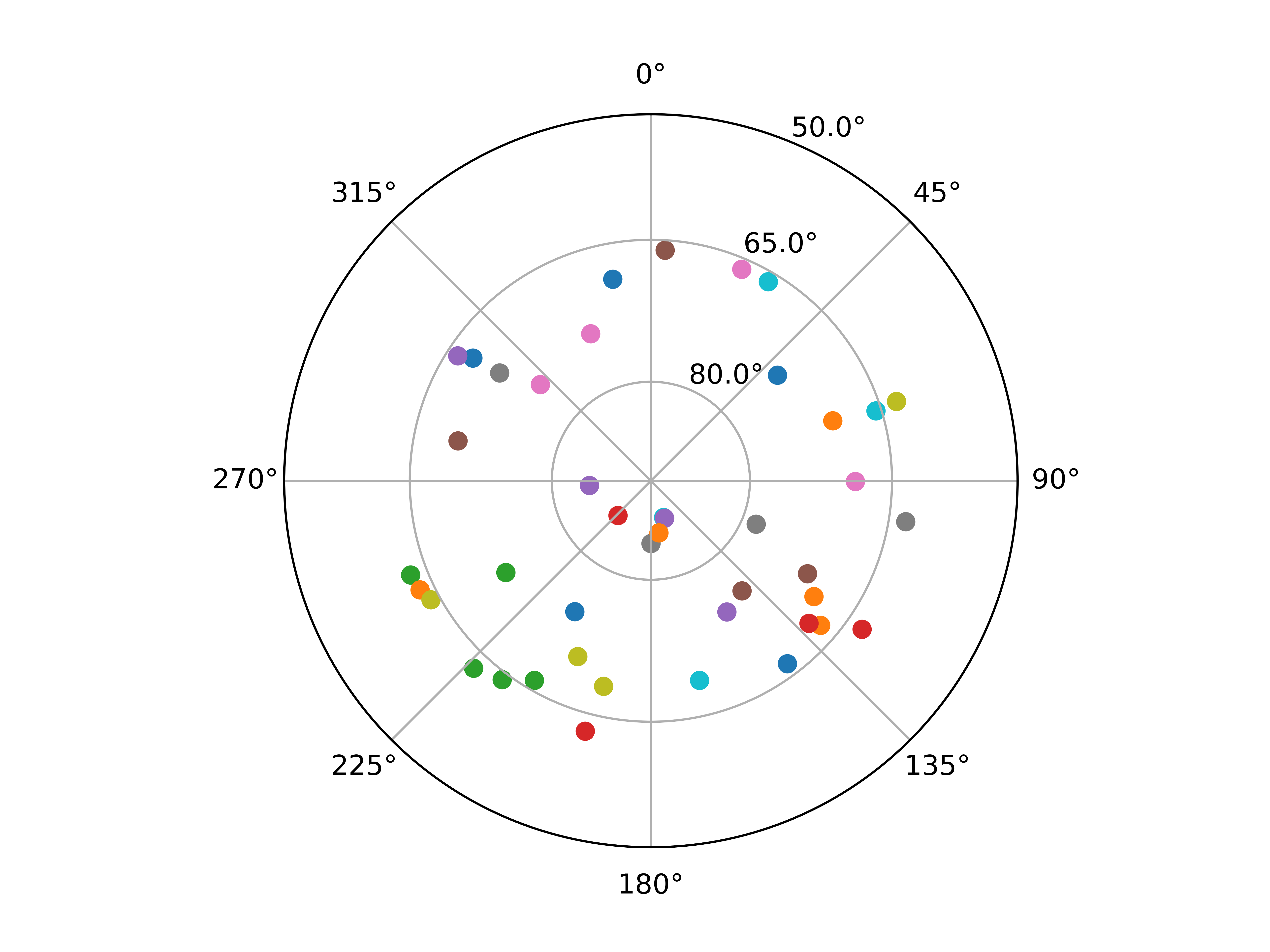}
    \caption{Viewing direction (OMA, NE)}
    \label{fig:Imaging properties of DFC2019 dataset-d}
  \end{subfigure}
  \caption{Imaging properties of DFC2019 dataset}
  \label{fig:Imaging properties of DFC2019 dataset}
\end{figure}

Satellite pairs have been chosen from the DFC2019\cite{saux_data_2019} track 3 dataset, a multiple-date satellite image data processing challenge. These include stereo images captured by the WorldView-3 satellite sensor, which has a spatial resolution of 0.3 meters. Additionally, airborne LiDAR data (spatial resolution is 0.5 meters) are provided as ground-truth geometry for DSM analyses. The DFC2019 challenge provides 107 tiles covering over 40 square kilometers collected in Jacksonville, Florida (JAX, FL), and Omaha, Nebraska (OMA, NE). The JAX area includes 53 tiles, each image is cropped in 2048x2048 pixels, covering about 600x600 meters, while the OMA area includes 54 tiles. The image coverage for each tile varies slightly due to the differences in footprints of the multi-date satellite images. In JAX, 24 images were collected between October 2014 and February 2016. In OMA, 43 images were collected between September 2014 and November 2015. The collecting time of each image is plotted in \cref{fig:Image collection time of DFC2019 dataset}. The variety of sun direction and viewing direction are visualized in \cref{fig:Imaging properties of DFC2019 dataset}.

\begin{figure}[hbt!]
  \centering
  \includegraphics[height=5.5cm]{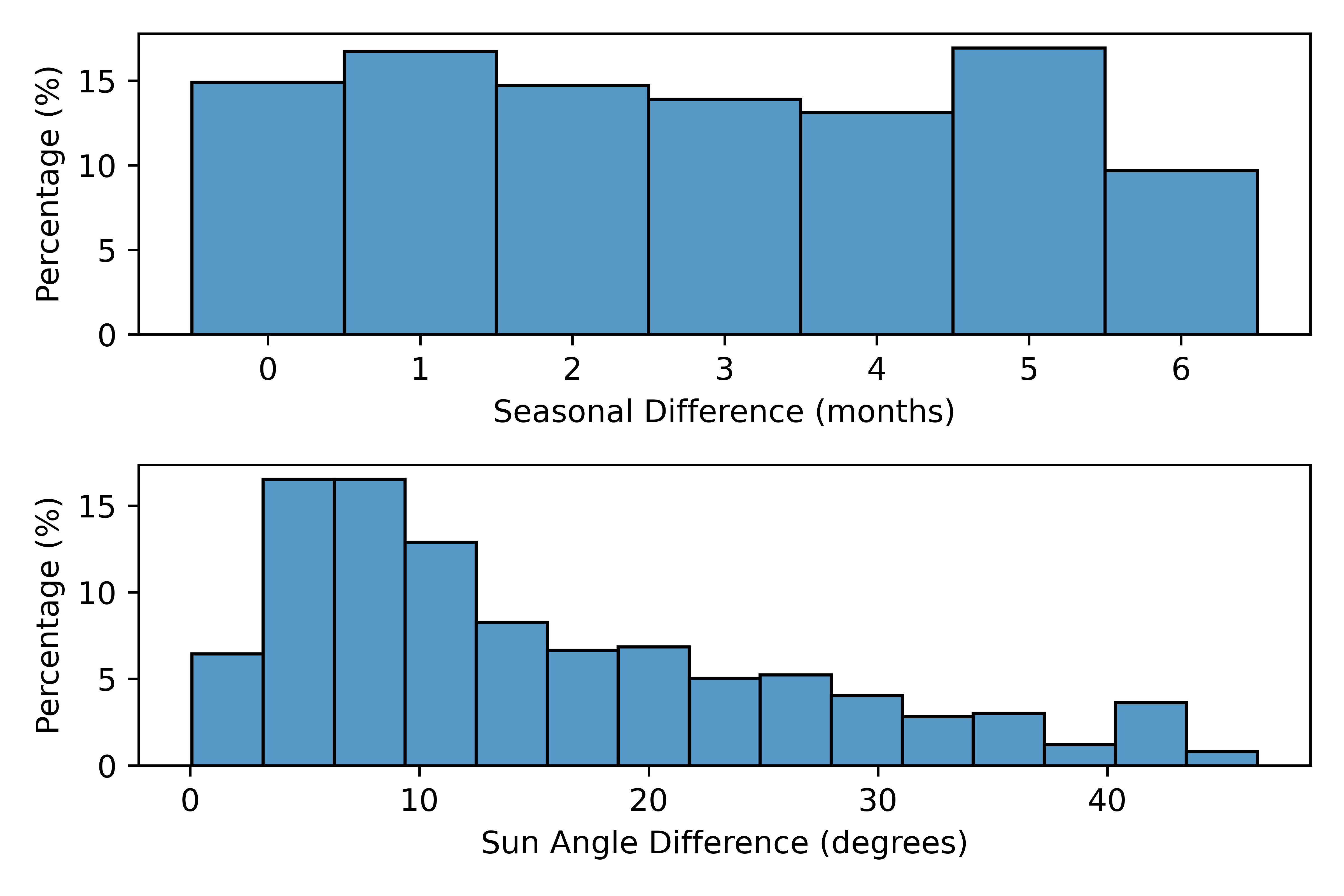}
  \caption{Characteristics of sampled pairs across all tiles
  }
  \label{fig:Characteristics of sampled pairs across all tiles}
\end{figure}

Employing the pair selection method outlined in \cref{sec:Methodology3.2}, we chose up to 5 stereo pairs from each tile, in total 496 sampled pairs in our evaluation. The statistics of the properties of sampled pairs are shown in \cref{fig:Characteristics of sampled pairs across all tiles}.

\begin{figure}[hbt!]
  \centering
  \begin{subfigure}{0.49\linewidth}
    \centering
    \includegraphics[height=5.5cm]{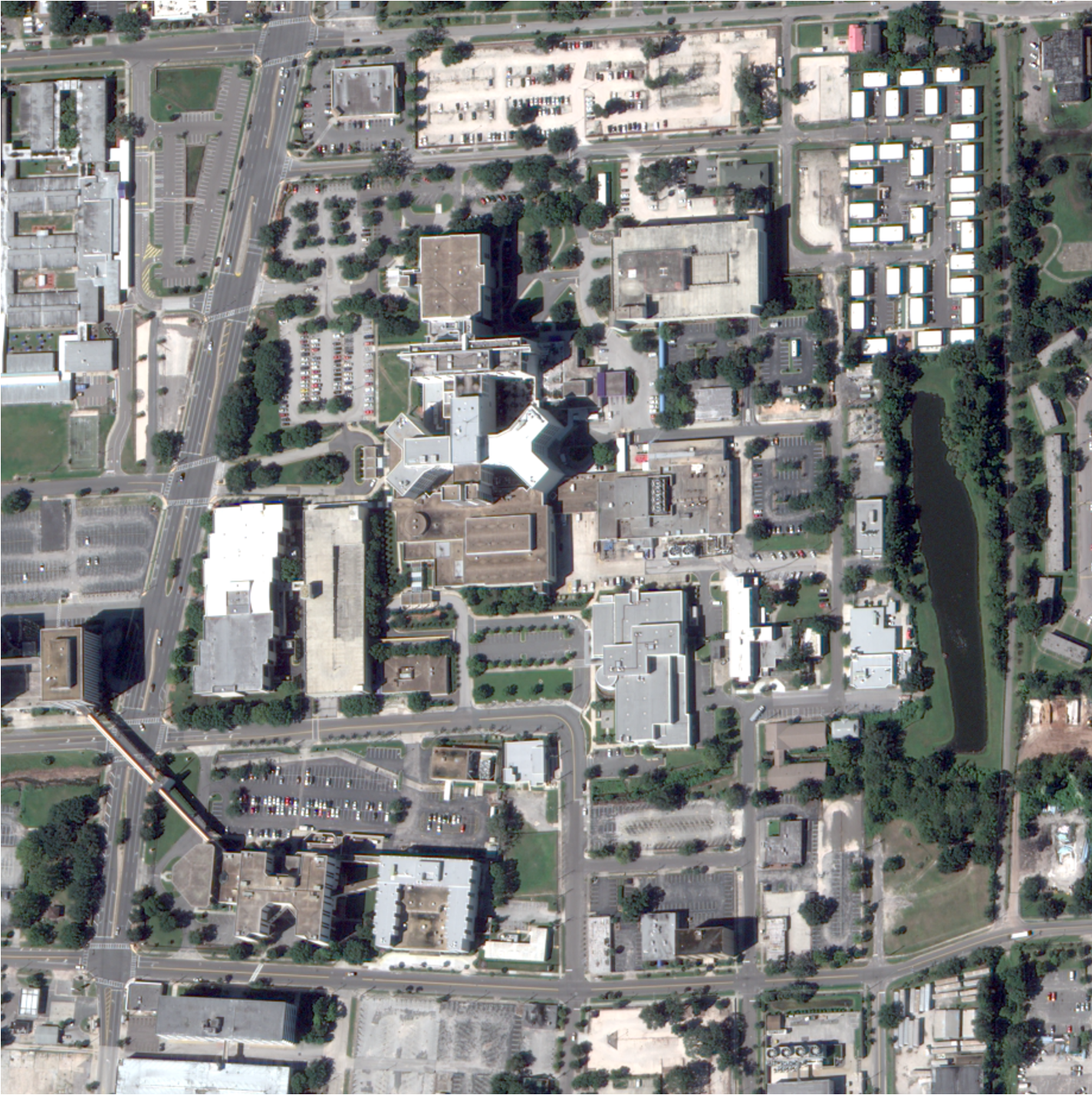}
    \caption{JAX, FL}
    \label{fig:Samples of the evaluation sites-a}
  \end{subfigure}
  \hfill
  \begin{subfigure}{0.49\linewidth}
    \centering
    \includegraphics[height=5.5cm]{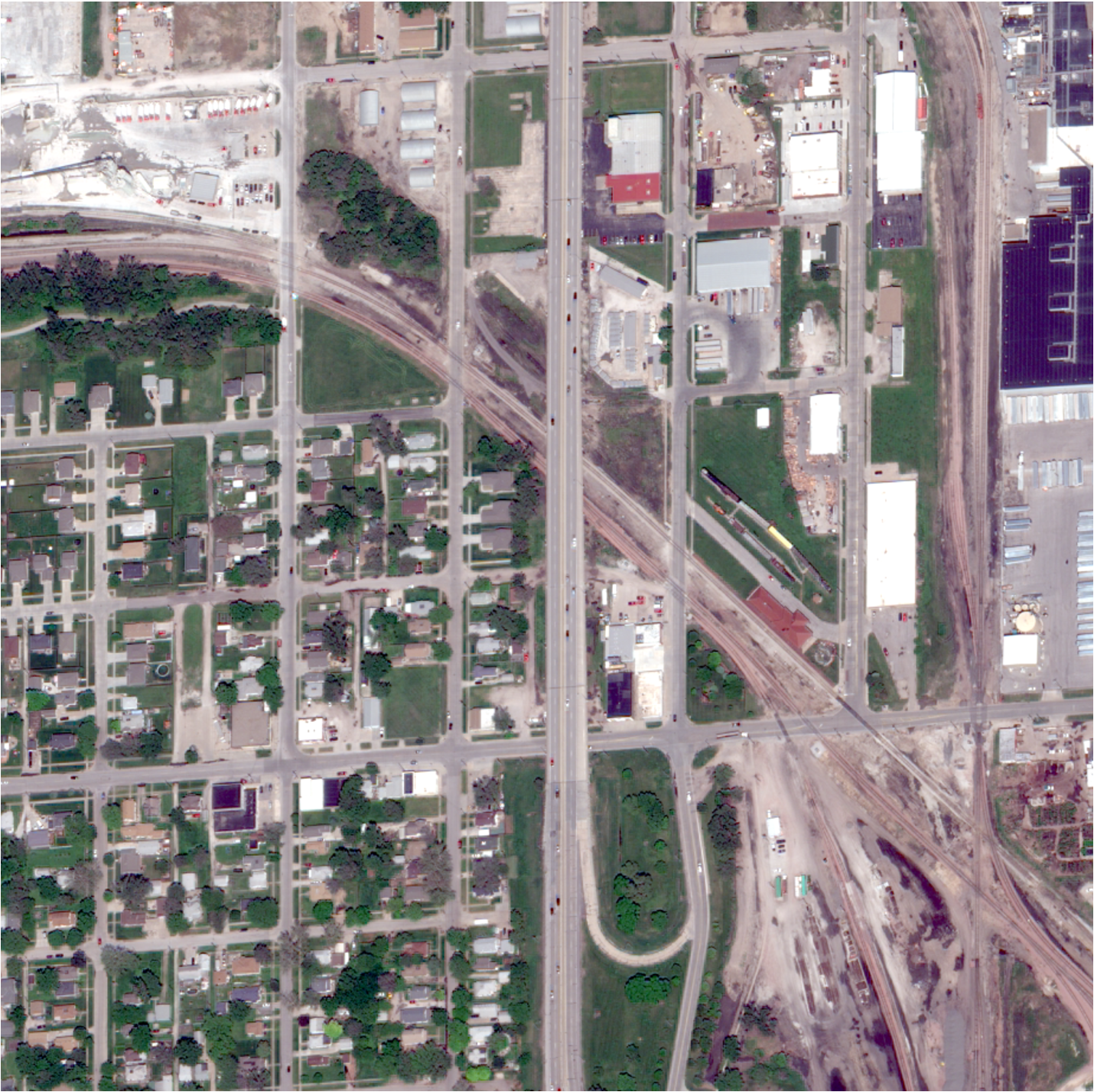}
    \caption{OMA, NE}
    \label{fig:Samples of the evaluation sites-b}
  \end{subfigure}
  \caption{Samples of the evaluation sites}
  \label{fig:Samples of the evaluation sites}
\end{figure}

\subsection{Analysis with Relative Orientation}
\label{sec:Experiments4.2}
\begin{figure}[hbt!]
  \centering
  \includegraphics[height=4.3cm]{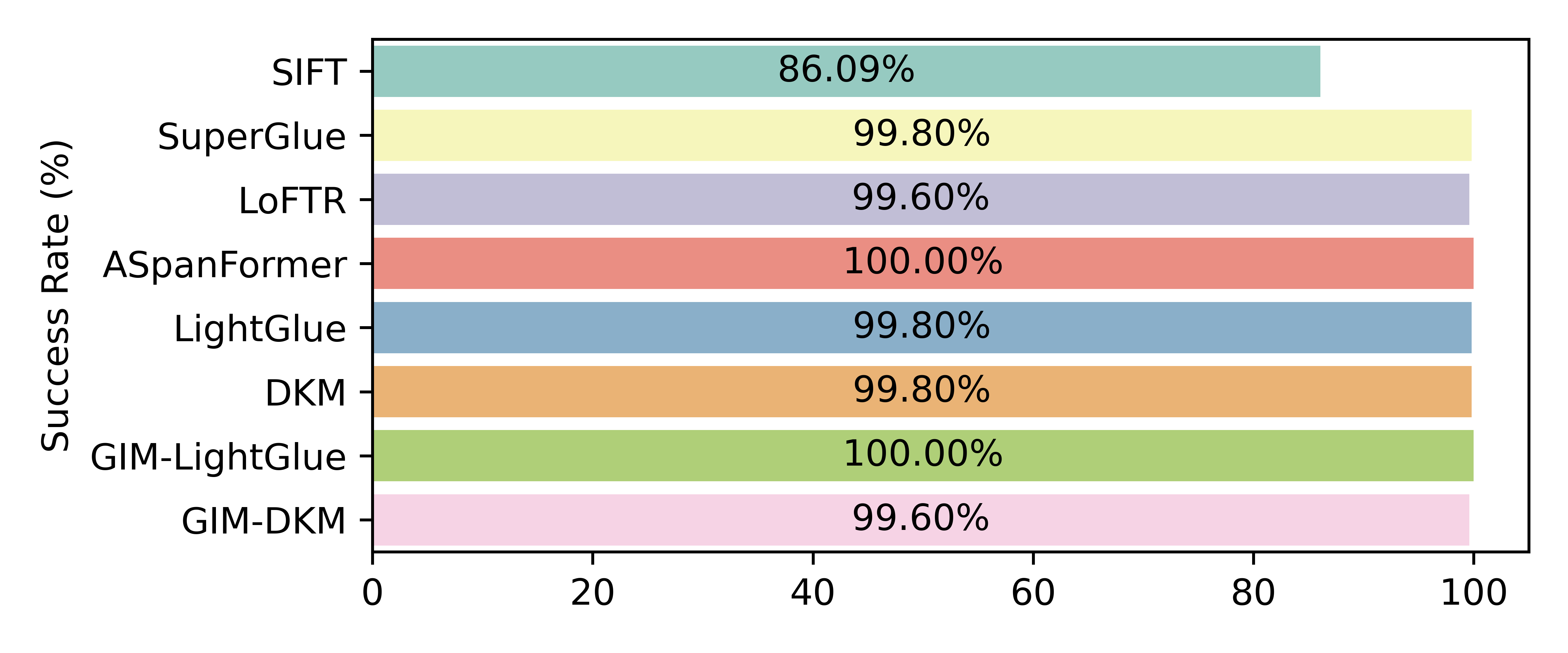}
  \caption{\emph{Success rates} of relative orientation}
  \label{fig:Success rates of relative orientation}
\end{figure}

The RPC-based relative orientation is evaluated in terms of inliers number and epipolar error. After RANSAC, if the number of inliers is less than 5, the relative orientation result is considered unreliable and therefore discarded. Based on this standard, the success rate of relative orientation is presented in \cref{fig:Success rates of relative orientation}.

\begin{figure}[hbt!]
  \centering
  \begin{subfigure}{0.45\linewidth}
    \centering
    \includegraphics[trim=20px 10px 20px 10px, clip, height=2.8cm]{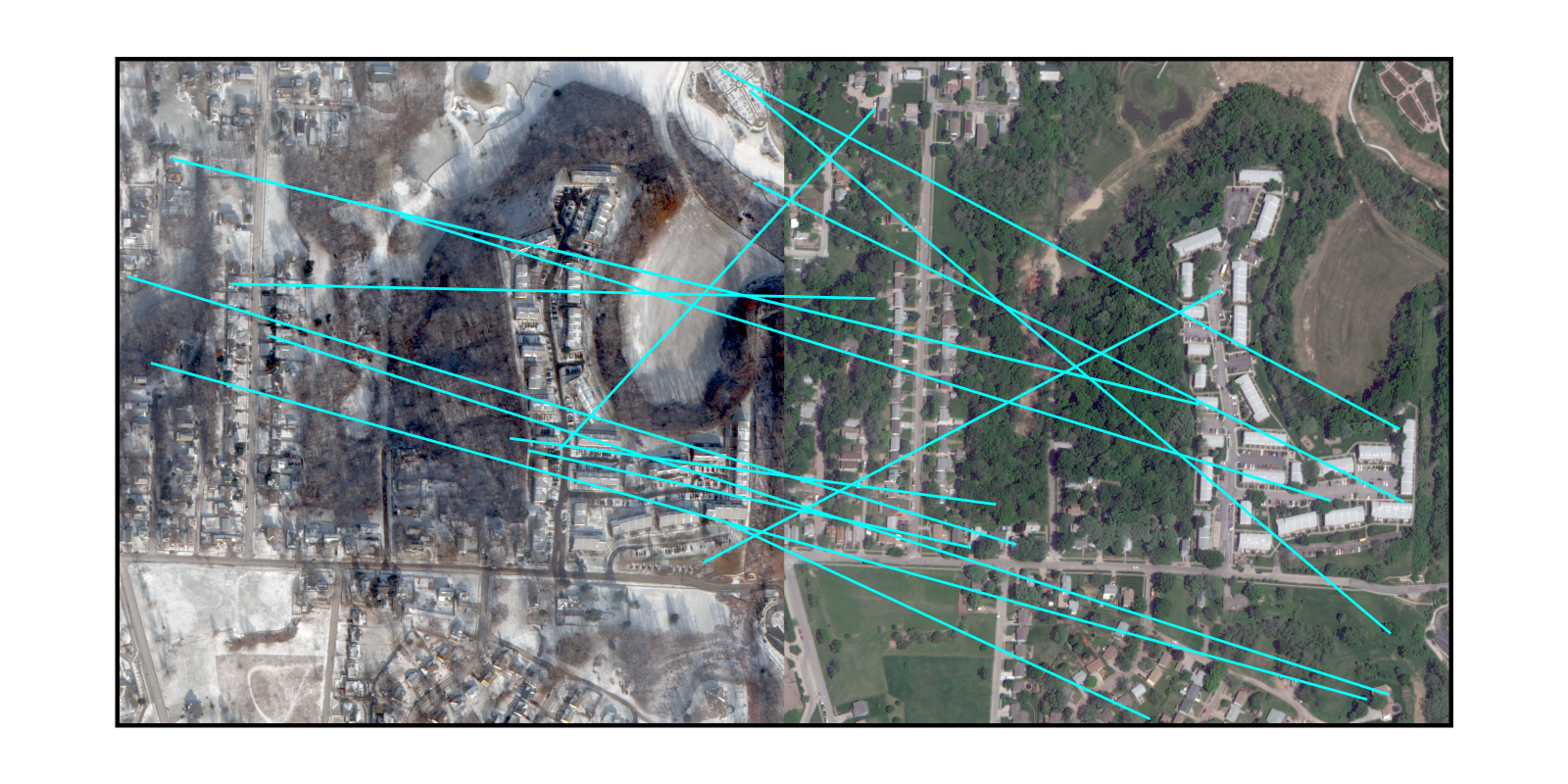}
    \caption{SIFT}
    \label{fig:sift_fail-a}
  \end{subfigure}
  \hfill
  \begin{subfigure}{0.45\linewidth}
    \centering
    \includegraphics[trim=20px 10px 20px 10px, clip, height=2.8cm]{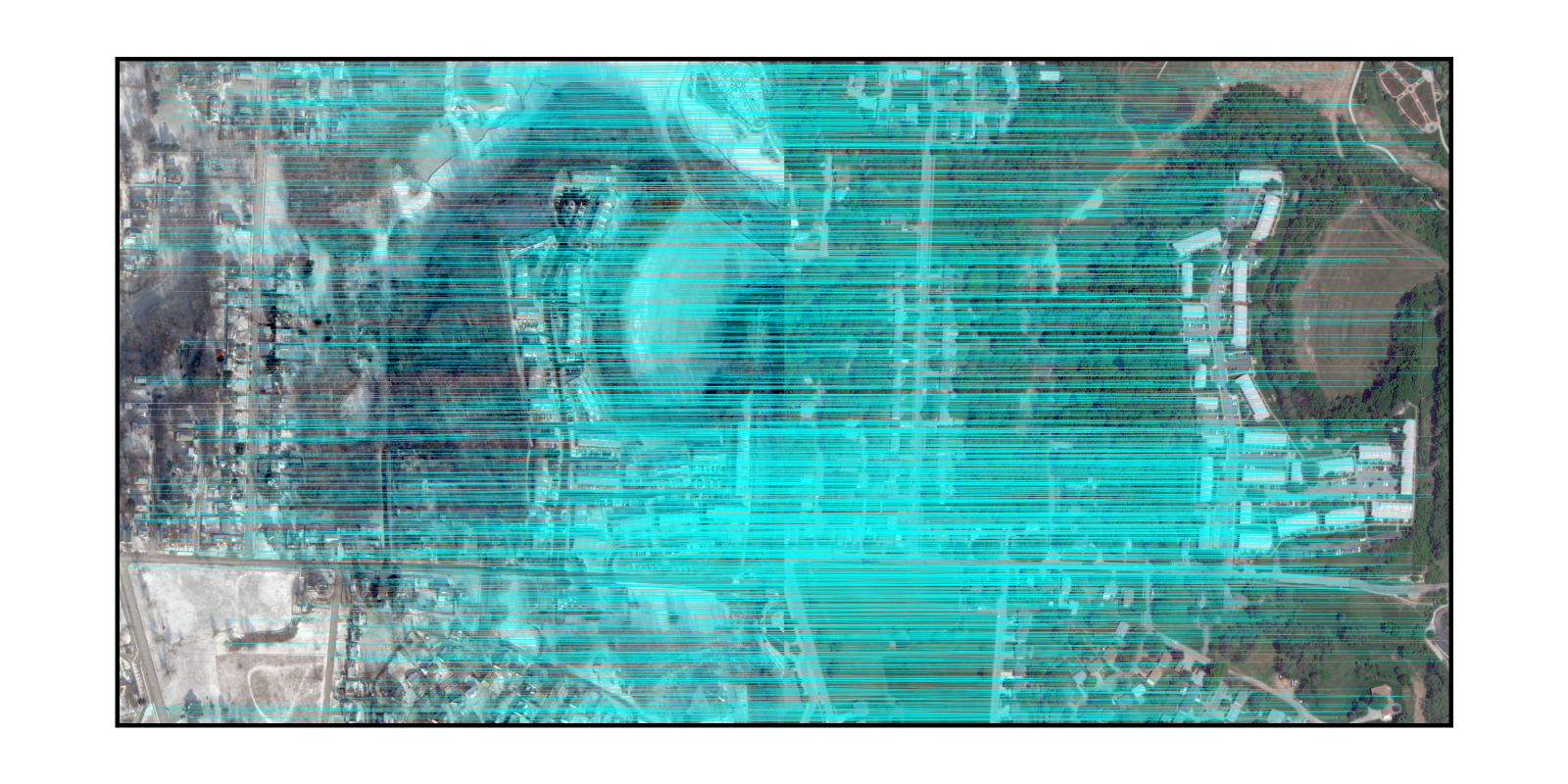}
    \caption{SuperGlue}
    \label{fig:sift_fail-b}
  \end{subfigure}
  \vfill
  \begin{subfigure}{0.45\linewidth}
    \centering
    \includegraphics[trim=20px 10px 20px 10px, clip, height=2.8cm]{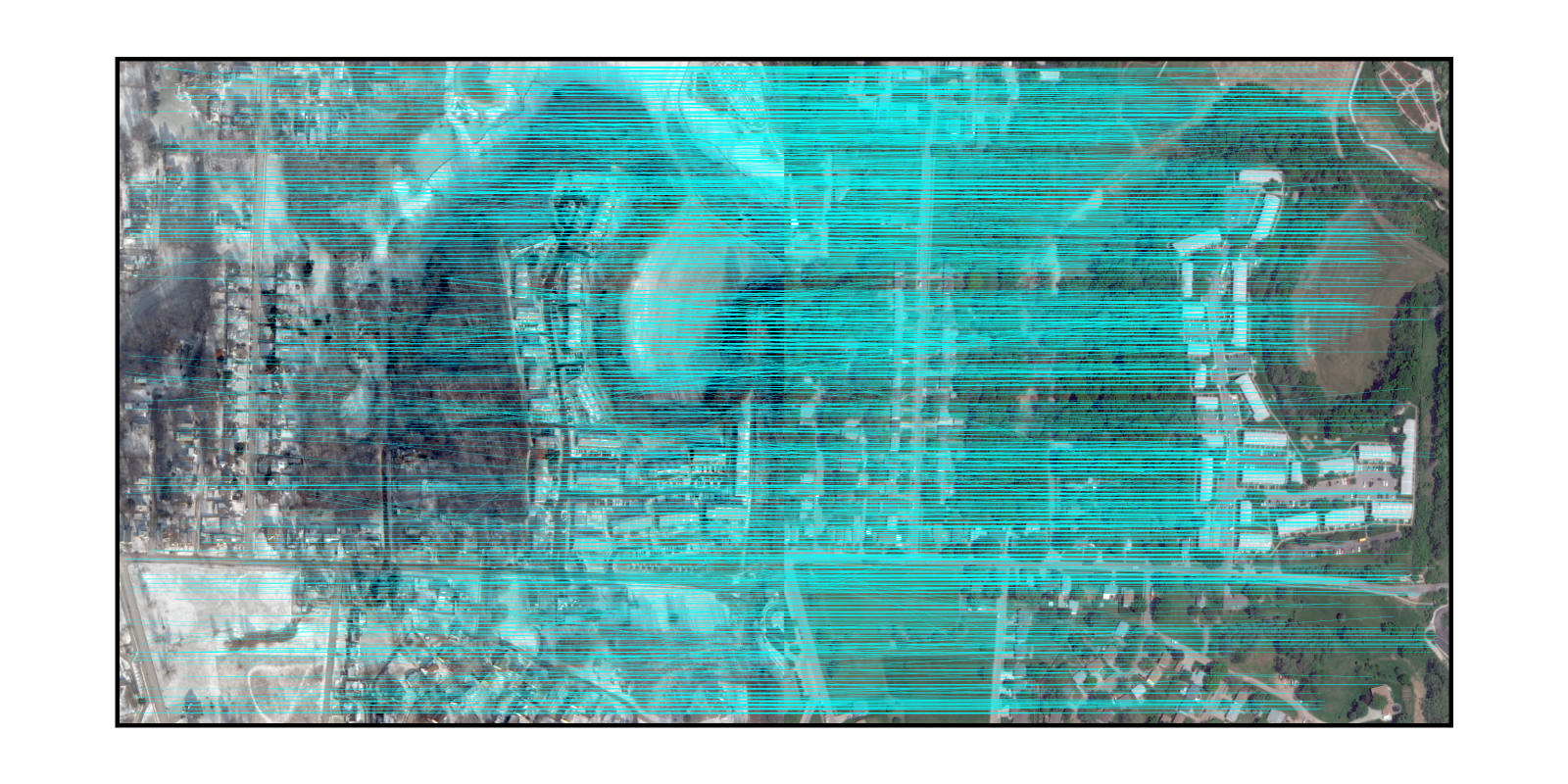}
    \caption{LoFTR}
    \label{fig:sift_fail-c}
  \end{subfigure}
  \hfill
  \begin{subfigure}{0.45\linewidth}
    \centering
    \includegraphics[trim=20px 10px 20px 10px, clip, height=2.8cm]{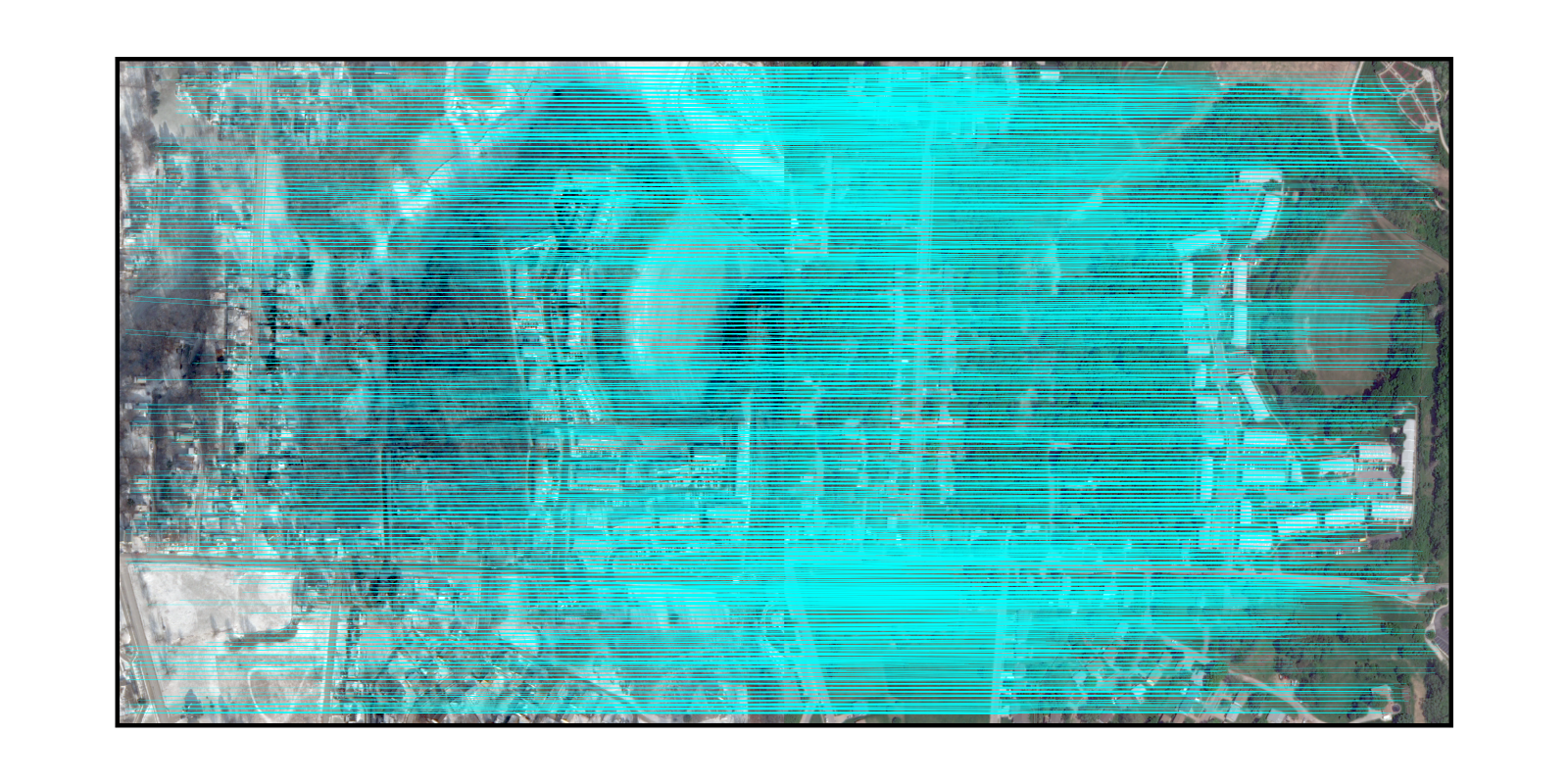}
    \caption{ASpanFormer}
    \label{fig:sift_fail-d}
  \end{subfigure}
  \vfill
  \begin{subfigure}{0.45\linewidth}
    \centering
    \includegraphics[trim=20px 10px 20px 10px, clip, height=2.8cm]{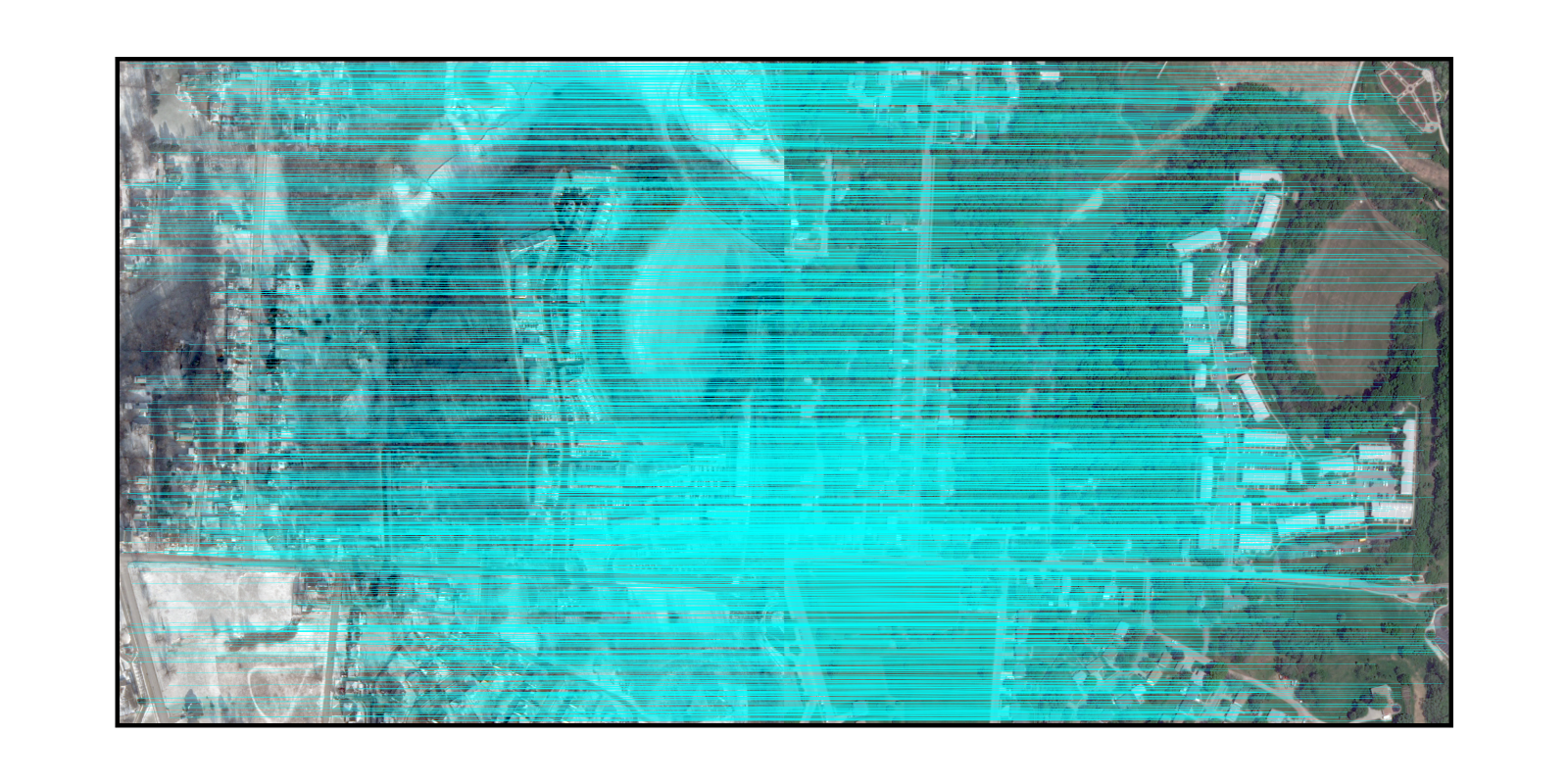}
    \caption{LightGlue}
    \label{fig:sift_fail-e}
  \end{subfigure}
  \hfill
  \begin{subfigure}{0.45\linewidth}
    \centering
    \includegraphics[trim=20px 10px 20px 10px, clip, height=2.8cm]{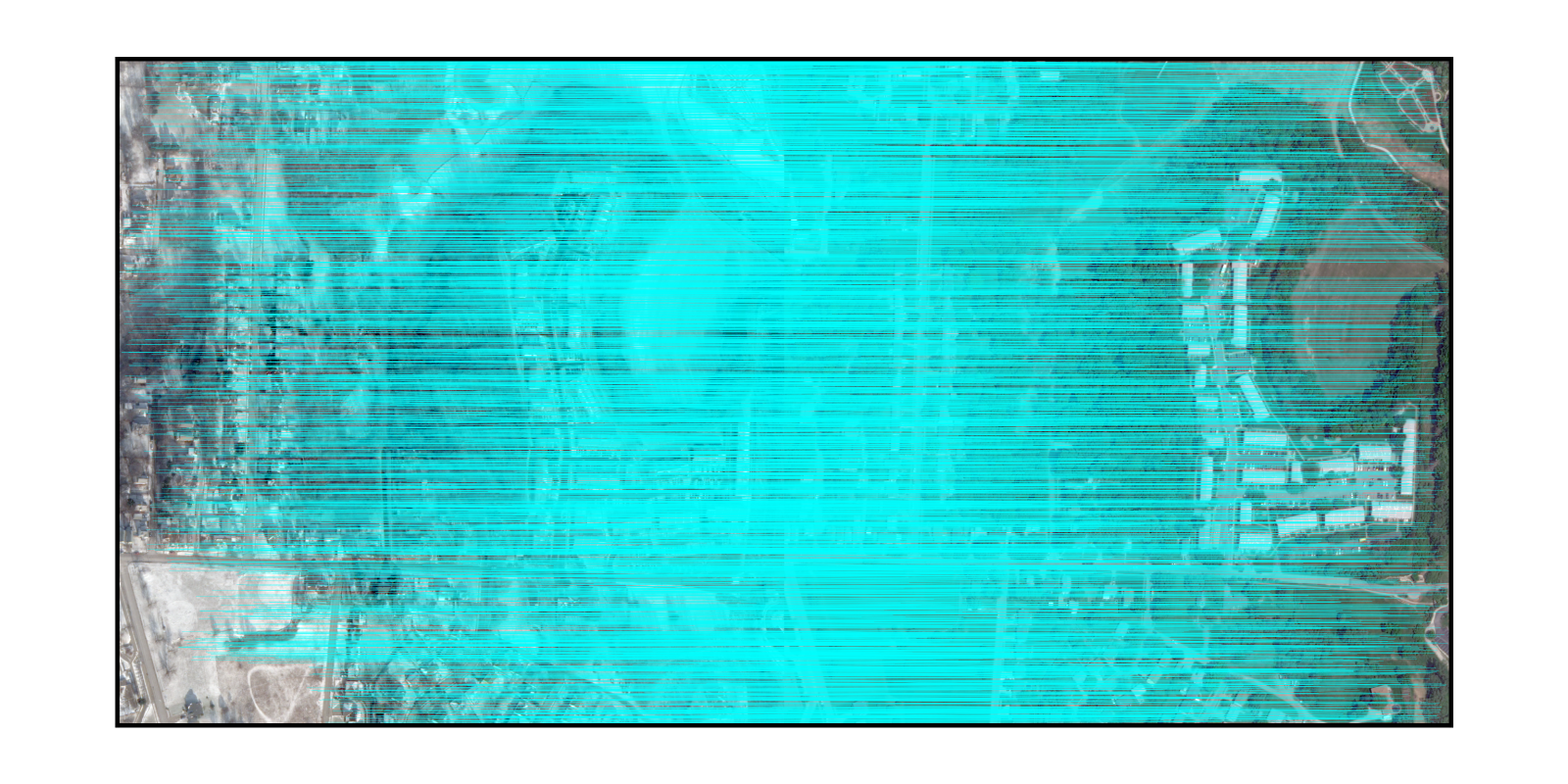}
    \caption{DKM}
    \label{fig:sift_fail-f}
  \end{subfigure}
  \vfill
  \begin{subfigure}{0.45\linewidth}
    \centering
    \includegraphics[trim=20px 10px 20px 10px, clip, height=2.8cm]{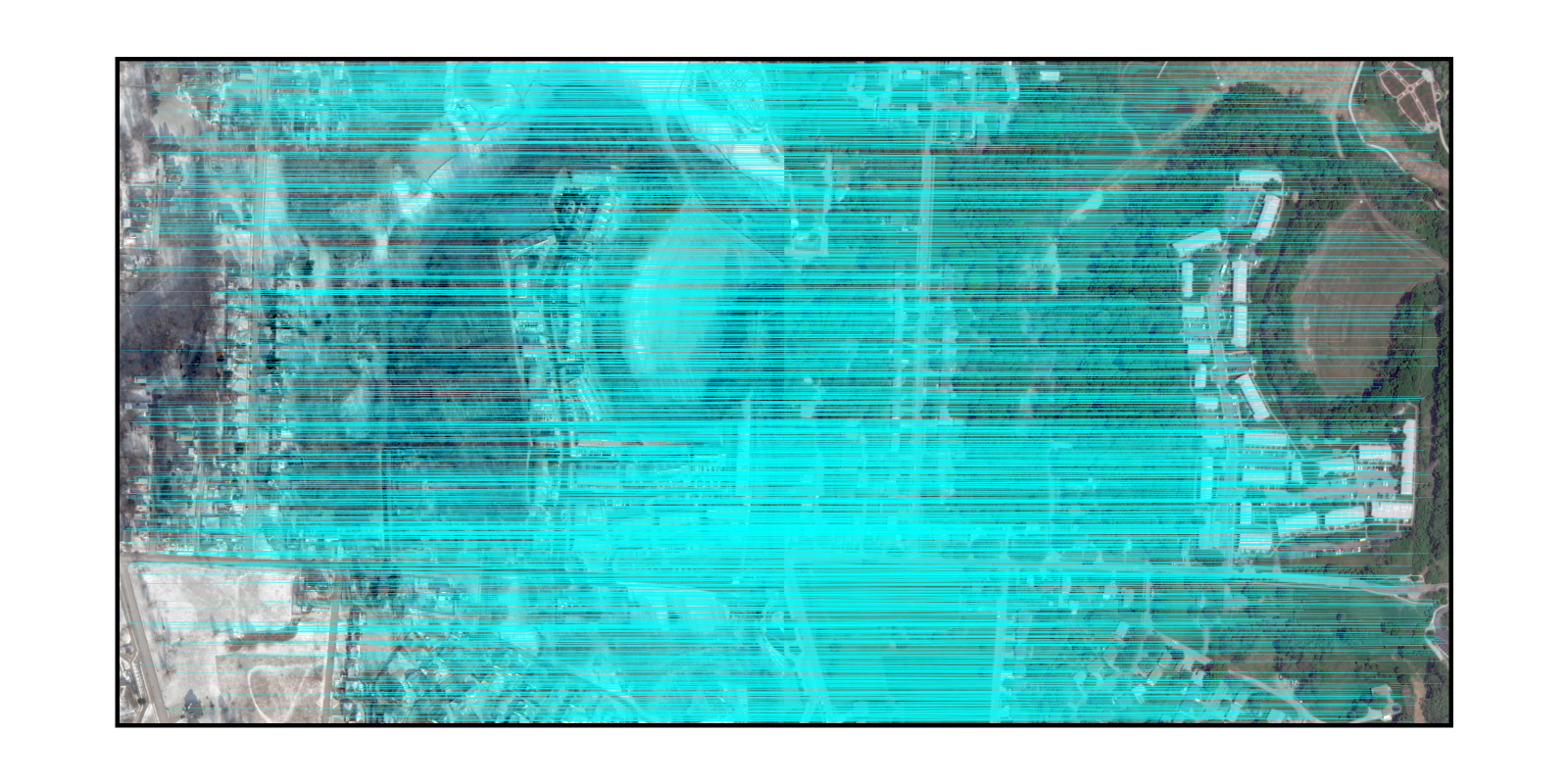}
    \caption{GIM-LightGlue}
    \label{fig:sift_fail-g}
  \end{subfigure}
  \hfill
  \begin{subfigure}{0.45\linewidth}
    \centering
    \includegraphics[trim=20px 10px 20px 10px, clip, height=2.8cm]{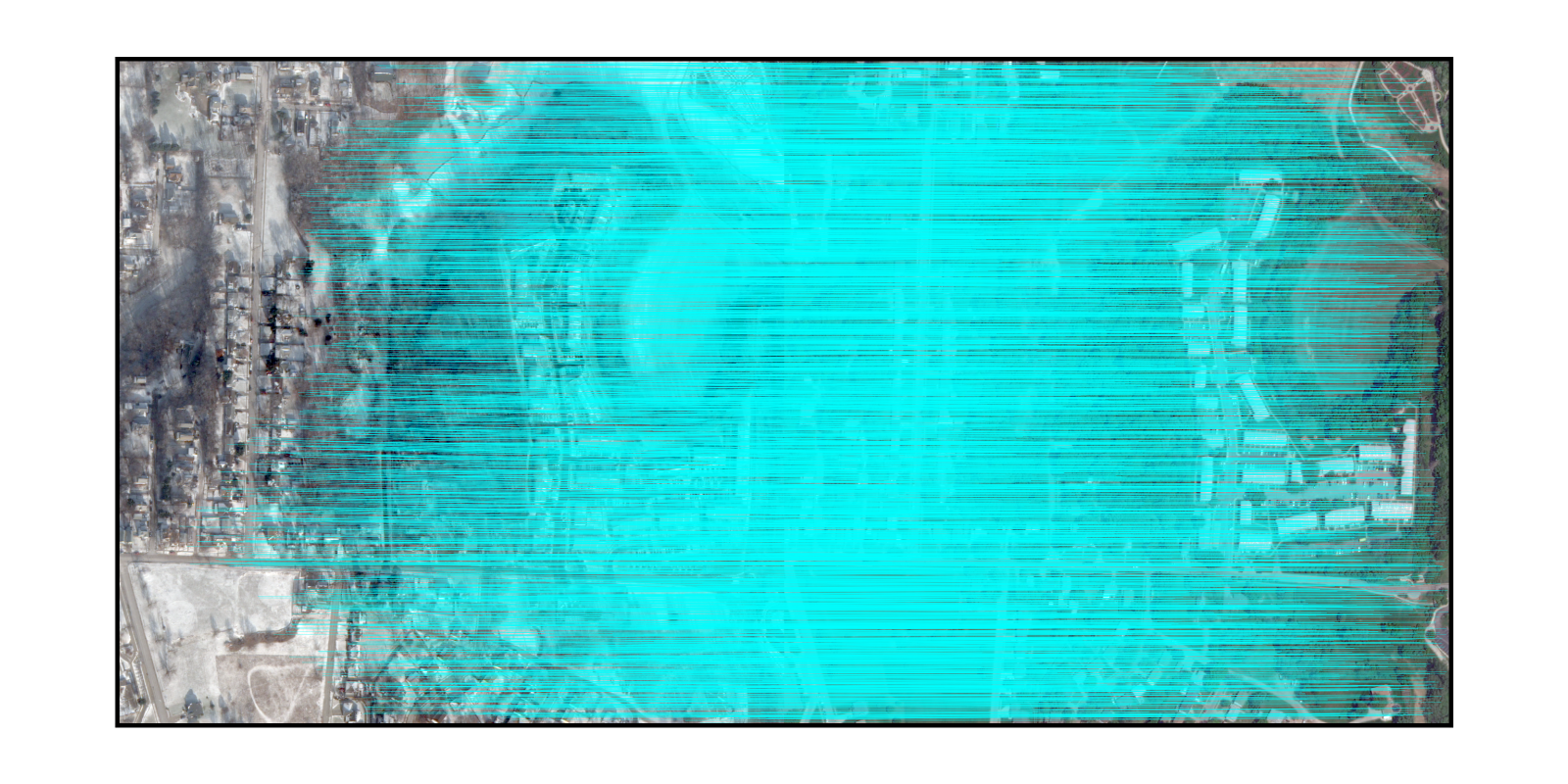}
    \caption{GIM-DKM}
    \label{fig:sift_fail-h}
  \end{subfigure}
  \caption{An example in OMA scene where  SIFT failed in the relative orientation due to too less inliers whereas other learning-based matcher successes.}
  \label{fig:sift_fail}
\end{figure}

An important finding is that SIFT matching shows significantly least success rate. This result is illustrated in \cref{fig:sift_fail}, where a pair of images and their matches are reported. The failure is attributed to significant texture changes caused by seasonal differences.

\begin{figure}[hbt!]
  \centering
  \includegraphics[height=4.3cm]{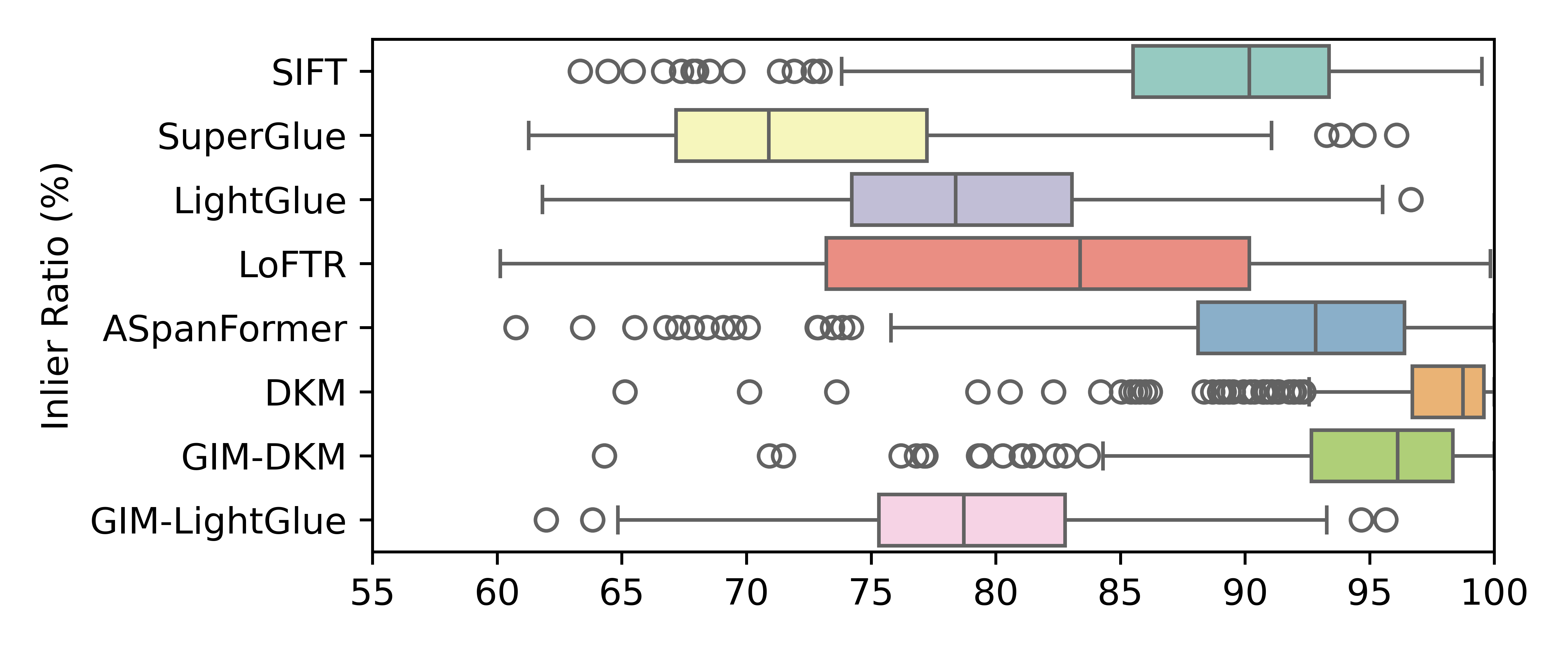}
  \caption{The box plot of \emph{inlier ratio} after relative orientation with RANSAC. Larger ratios are preferred.
  }
  \label{fig:inlier ratio}
\end{figure}

A further examination of the inlier ratio statistics of feature matching methods, as shown in \cref{fig:inlier ratio}, shows that learning-based methods with key point detectors present less inlier ratio than detector-free methods. Particularly DKM, constantly provides correspondences greater than 95\% inlier ratio. An interesting finding is that SIFT outperforms learning-based methods with key point detectors like SuperGlue, LightGlue, and GIM-LightGlue in terms of inlier ratio.

\begin{figure}[hbt!]
  \centering
  \includegraphics[height=4.3cm]{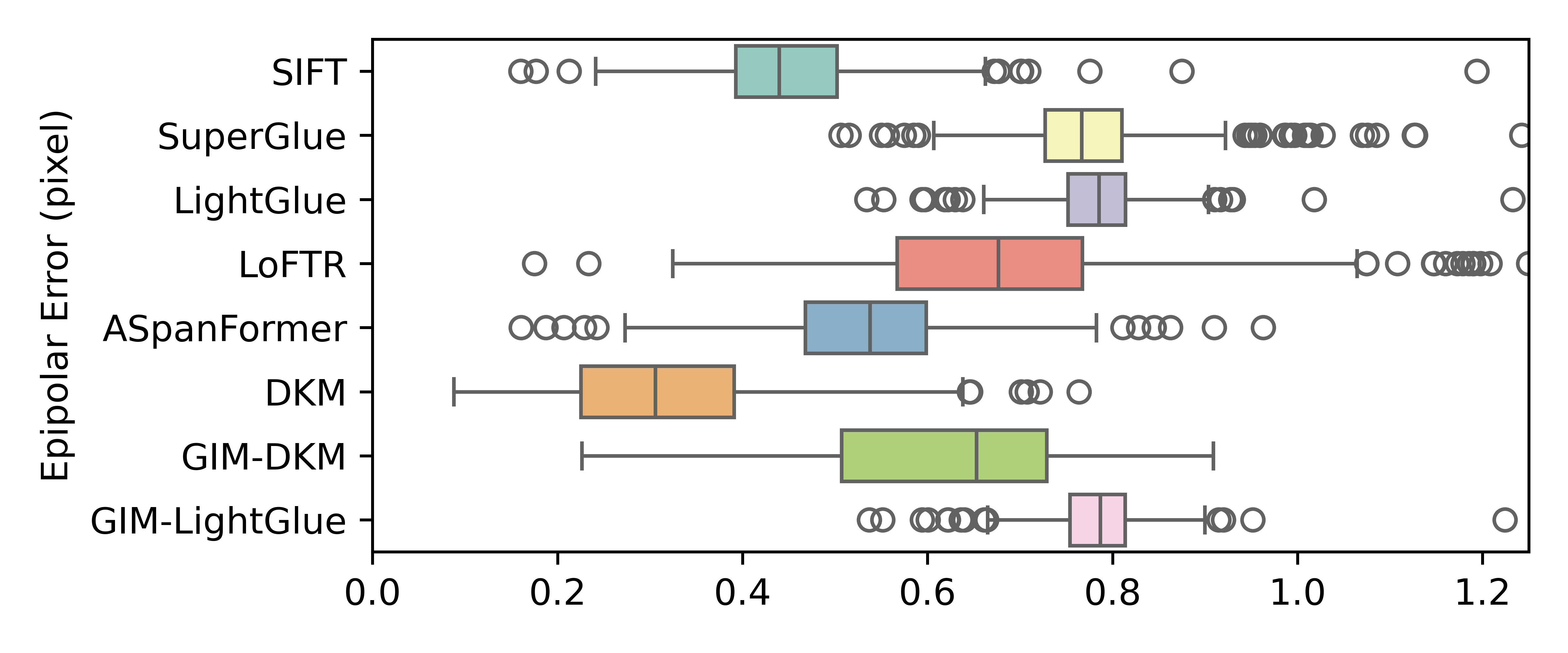}
  \caption{The box plot \emph{epipolar error} after relative orientation with RANSAC. Smaller errors are preferred.
  }
  \label{fig:epipolar error}
\end{figure}

When evaluating the epipolar error, as depicted in \cref{fig:epipolar error}, (GIM-)DKM, ASpanFormer and LoFTR inliers demonstrate a smaller epipolar error than SuperGlue, LightGlue and GIM-LightGlue methods. Compared with learning-based method, SIFT is at 2\textsuperscript{nd} place in term of epipolar error, which is competitive if compared to those state-of-the-art learning-based methods. The larger epipolar error of SuperGlue/LightGlue could be explained with the matches from SuperPoint that are extracted at pixel level, while SIFT extracts keypoints with sub-pixel accuracy.  

\subsection{Analysis with Dense Stereo Matching}
\label{sec:Experiments4.3}

\begin{figure}[hbt!]
  \centering
  \begin{subfigure}{0.22\linewidth}
    \centering
    \includegraphics[height=3cm]{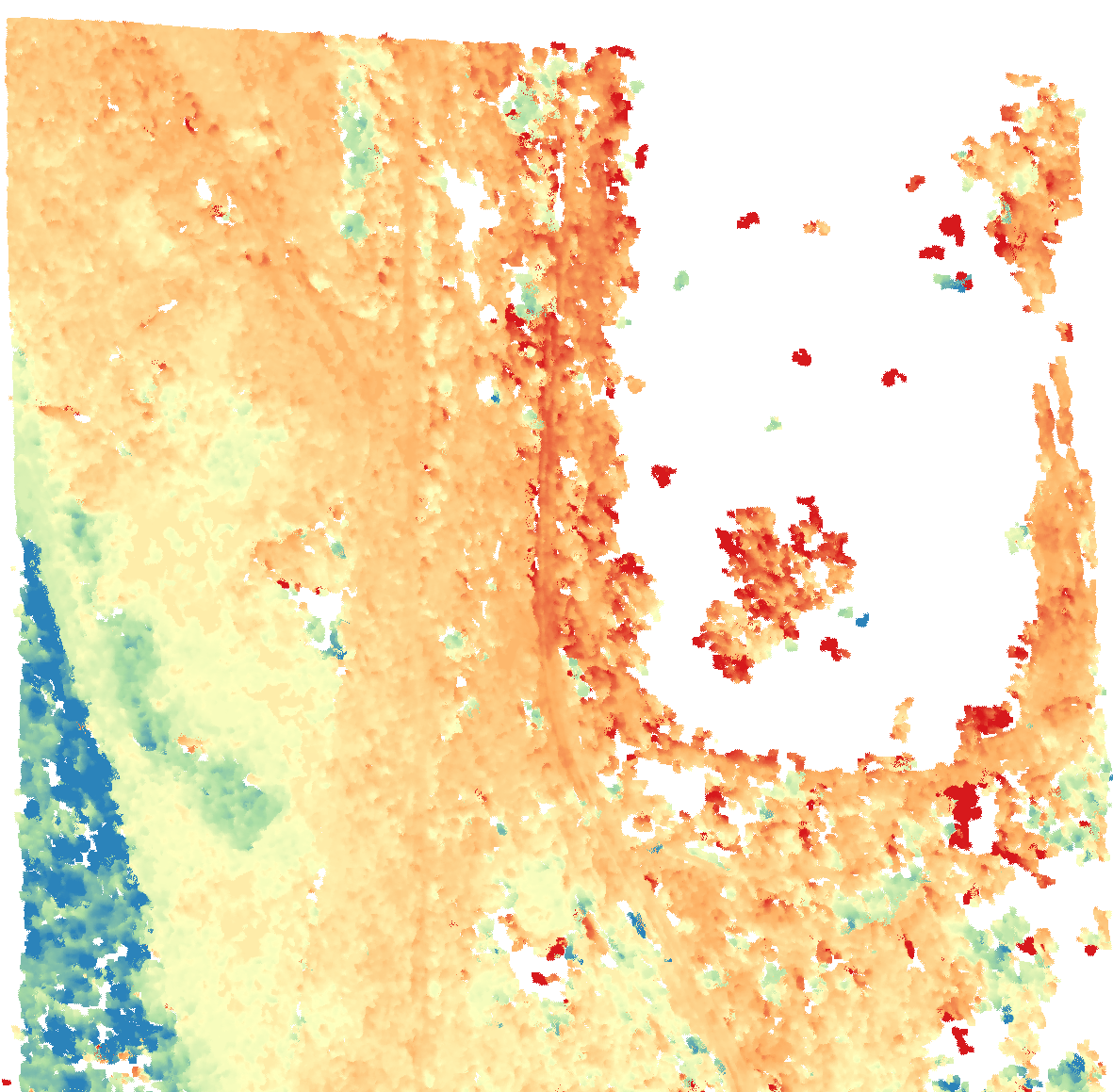}
    \caption{SIFT}
    \label{fig:Comparison between DSM using relative orientation from matchers-a}
  \end{subfigure}
  \hfill
  \begin{subfigure}{0.22\linewidth}
    \centering
    \includegraphics[height=3cm]{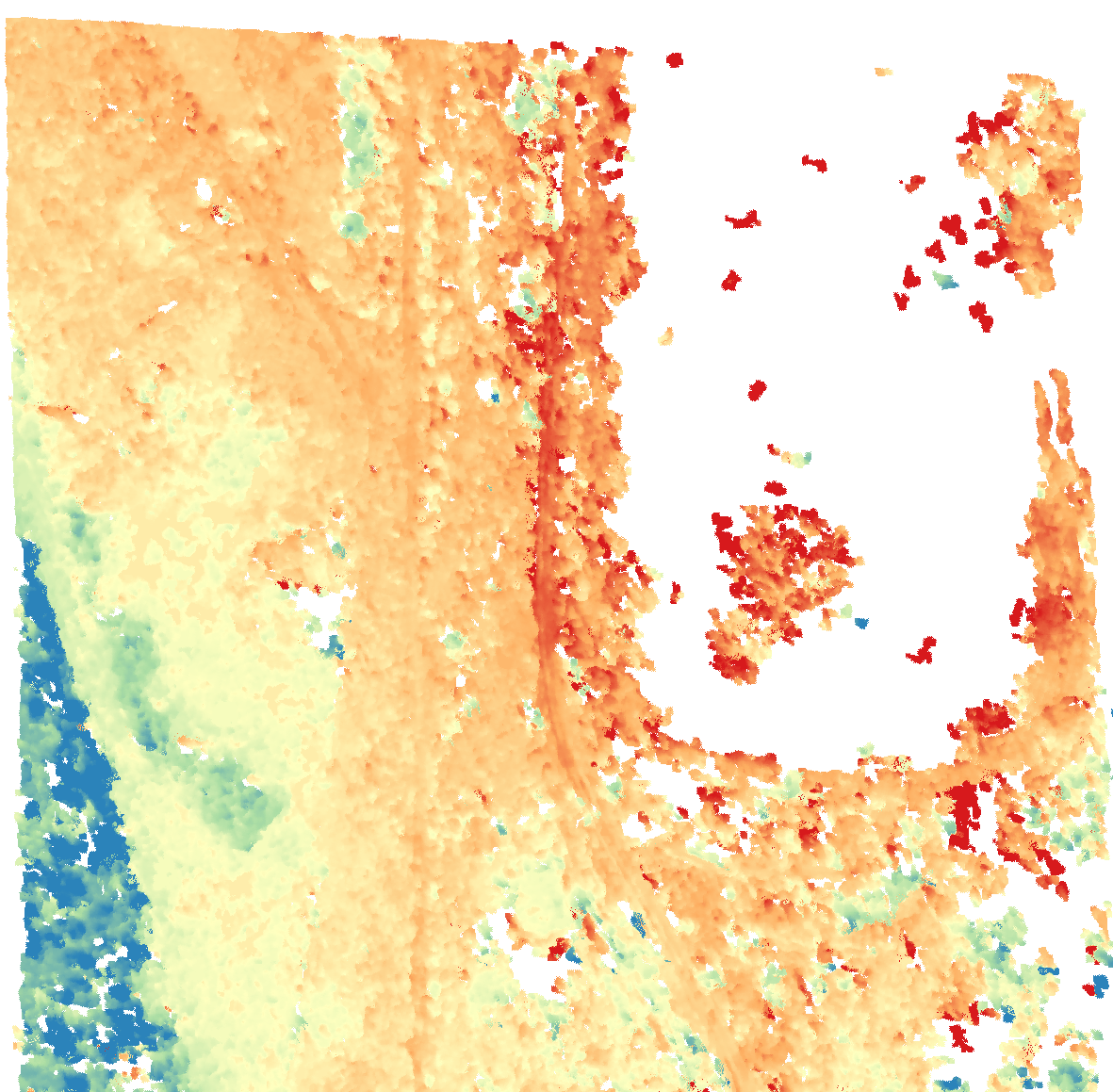}
    \caption{SuperGlue}
    \label{fig:Comparison between DSM using relative orientation from matchers-b}
  \end{subfigure}
  \hfill
  \begin{subfigure}{0.22\linewidth}
    \centering
    \includegraphics[height=3cm]{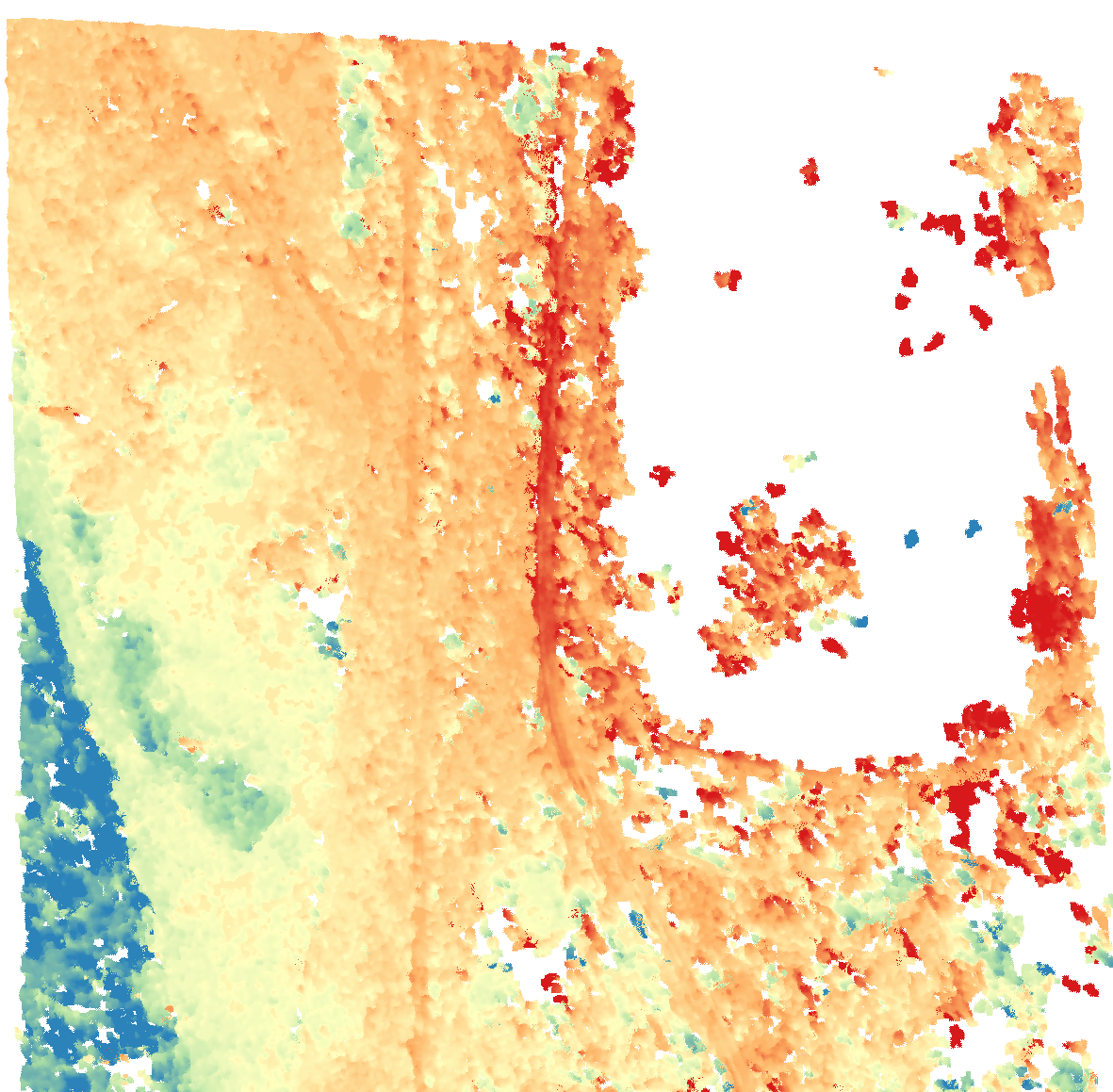}
    \caption{LightGlue}
    \label{fig:Comparison between DSM using relative orientation from matchers-c}
  \end{subfigure}
  \hfill
  \begin{subfigure}{0.22\linewidth}
    \centering
    \includegraphics[height=3cm]{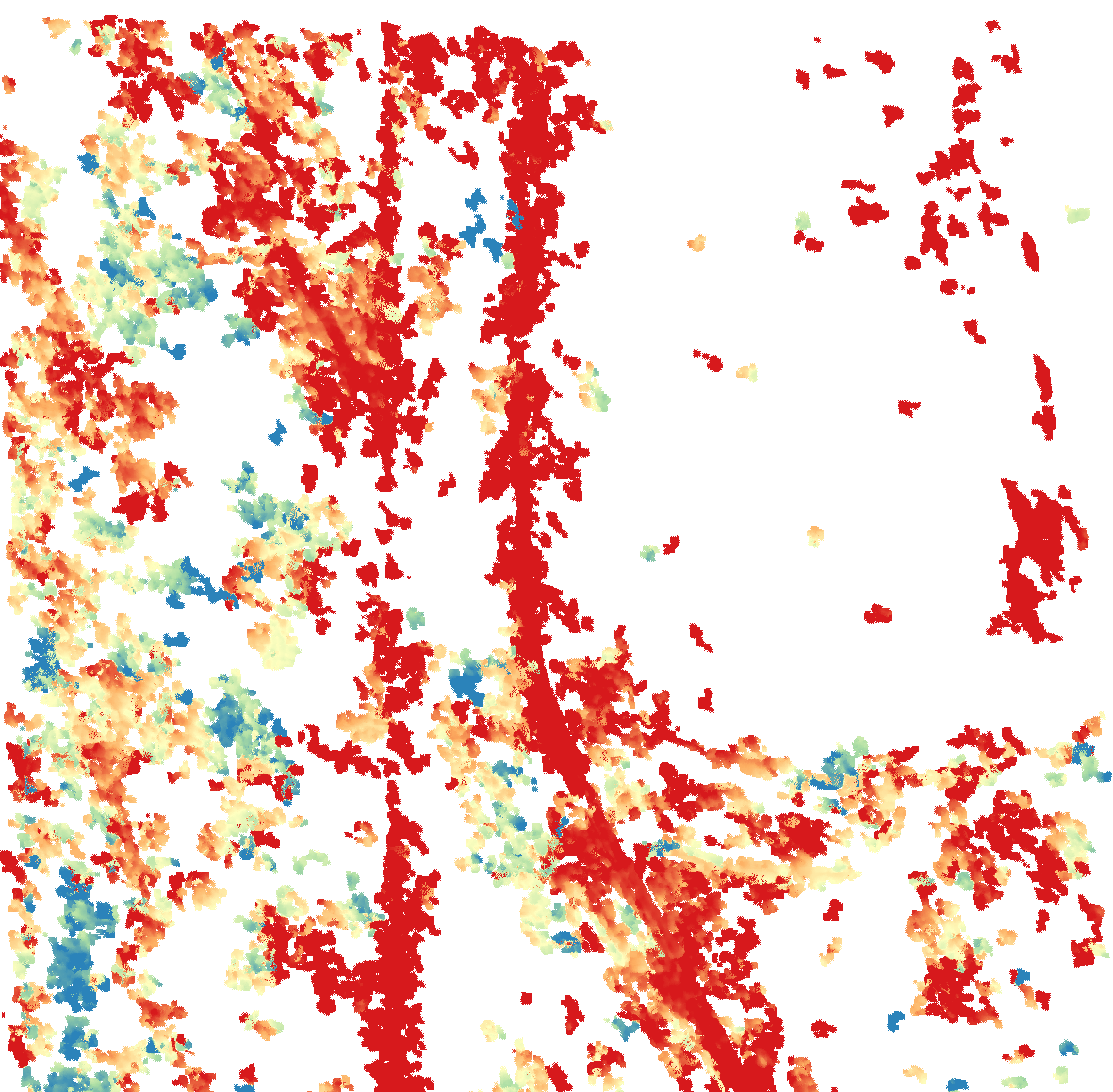}
    \caption{LoFTR}
    \label{fig:Comparison between DSM using relative orientation from matchers-d}
  \end{subfigure}
  \vfill
  \begin{subfigure}{0.22\linewidth}
    \centering
    \includegraphics[height=3cm]{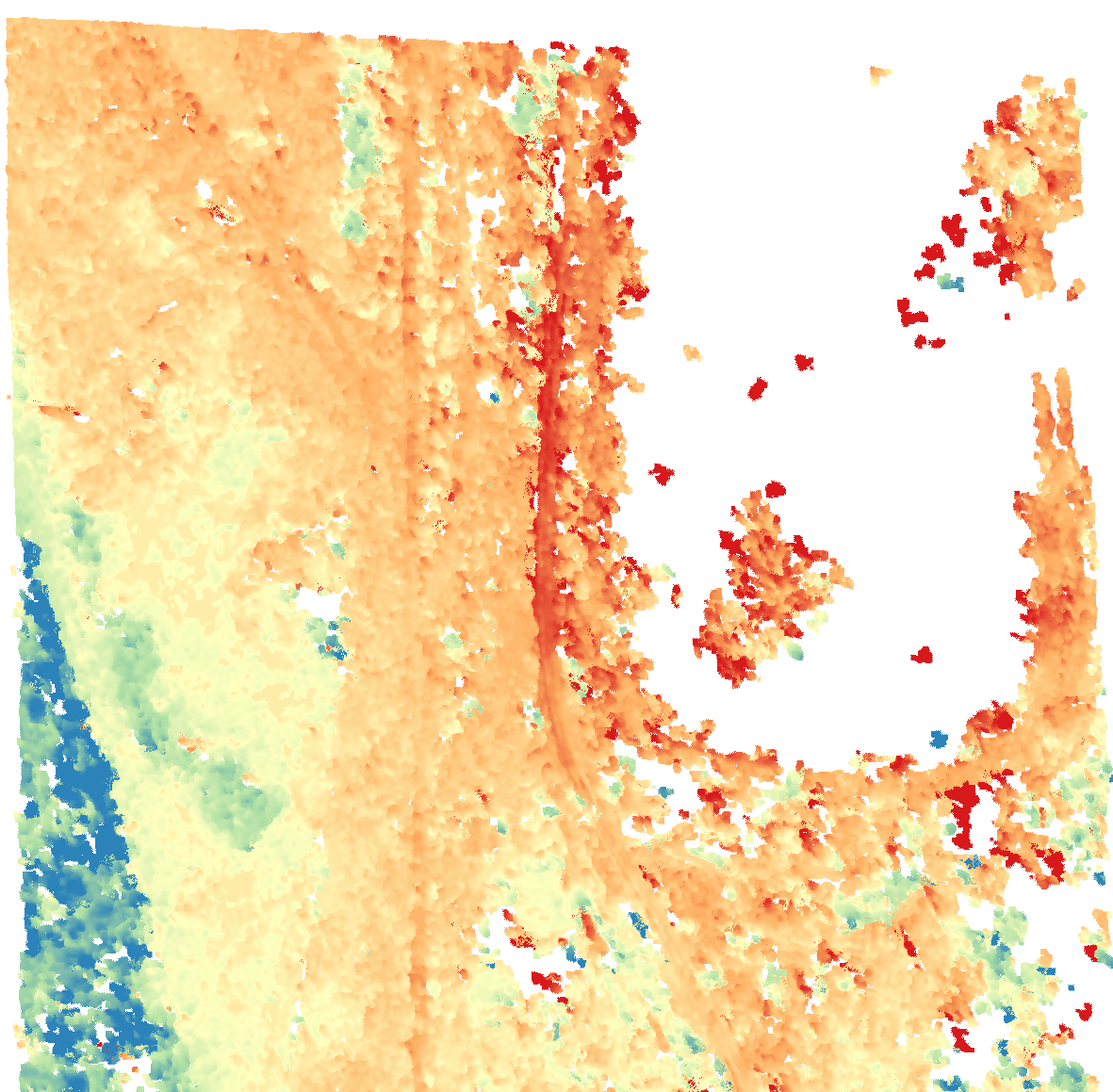}
    \caption{ASpanFormer}
    \label{fig:Comparison between DSM using relative orientation from matchers-e}
  \end{subfigure}
  \hfill
  \begin{subfigure}{0.22\linewidth}
    \centering
    \includegraphics[height=3cm]{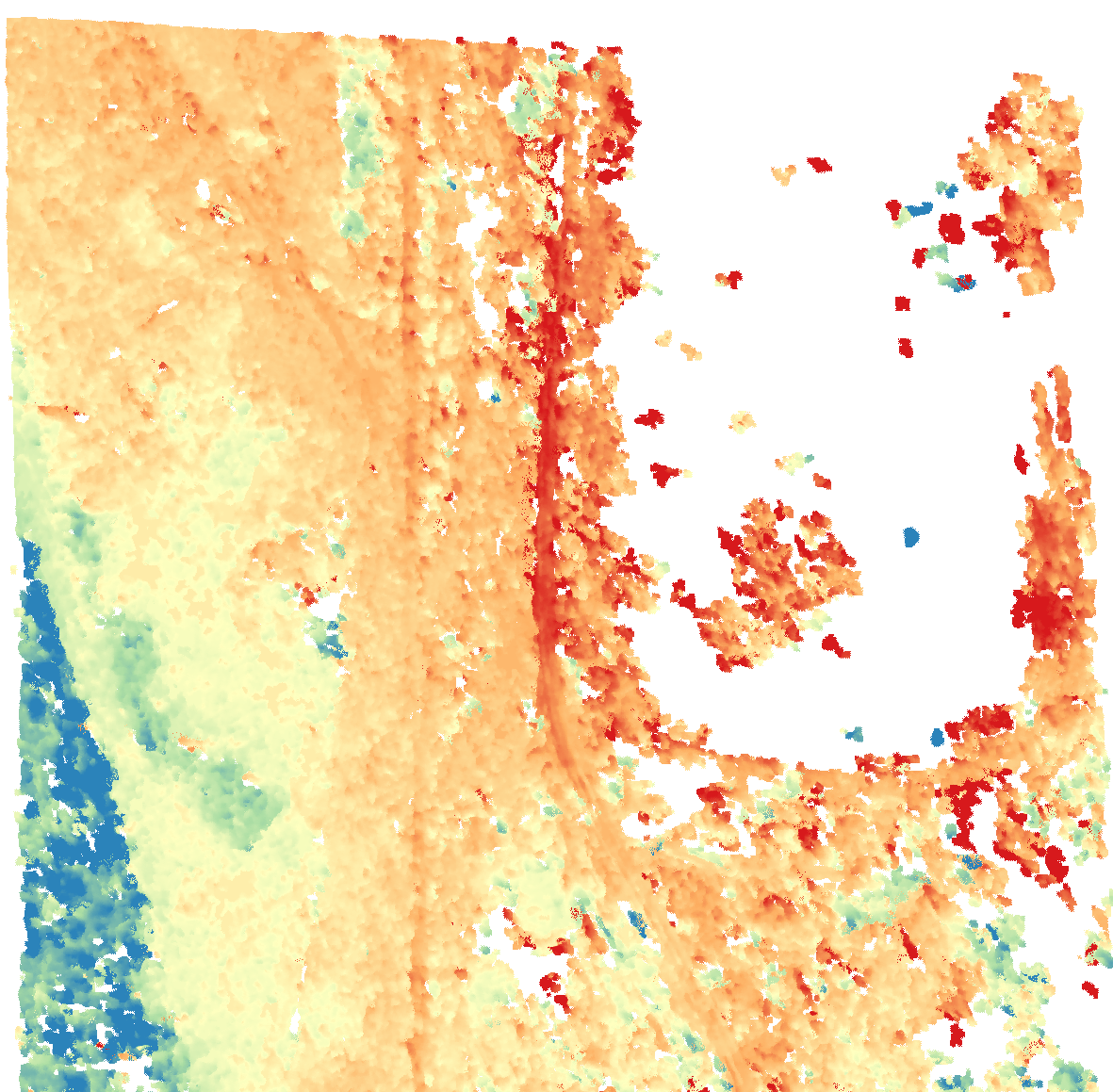}
    \caption{DKM}
    \label{fig:Comparison between DSM using relative orientation from matchers-f}
  \end{subfigure}
  \hfill
  \begin{subfigure}{0.22\linewidth}
    \centering
    \includegraphics[height=3cm]{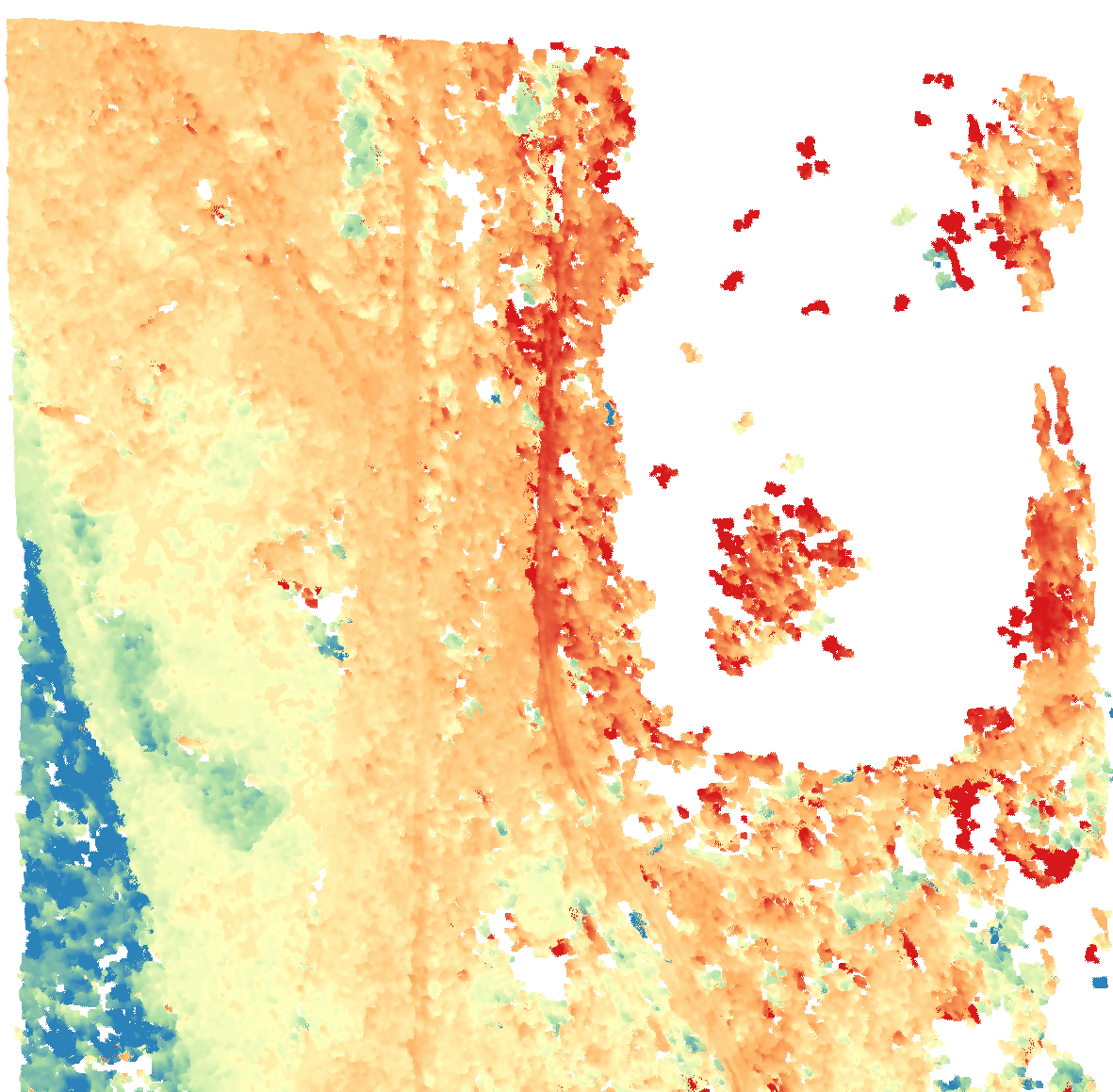}
    \caption{GIM-LightGlue}
    \label{fig:Comparison between DSM using relative orientation from matchers-g}
  \end{subfigure}
  \hfill
  \begin{subfigure}{0.22\linewidth}
    \centering
    \includegraphics[height=3cm]{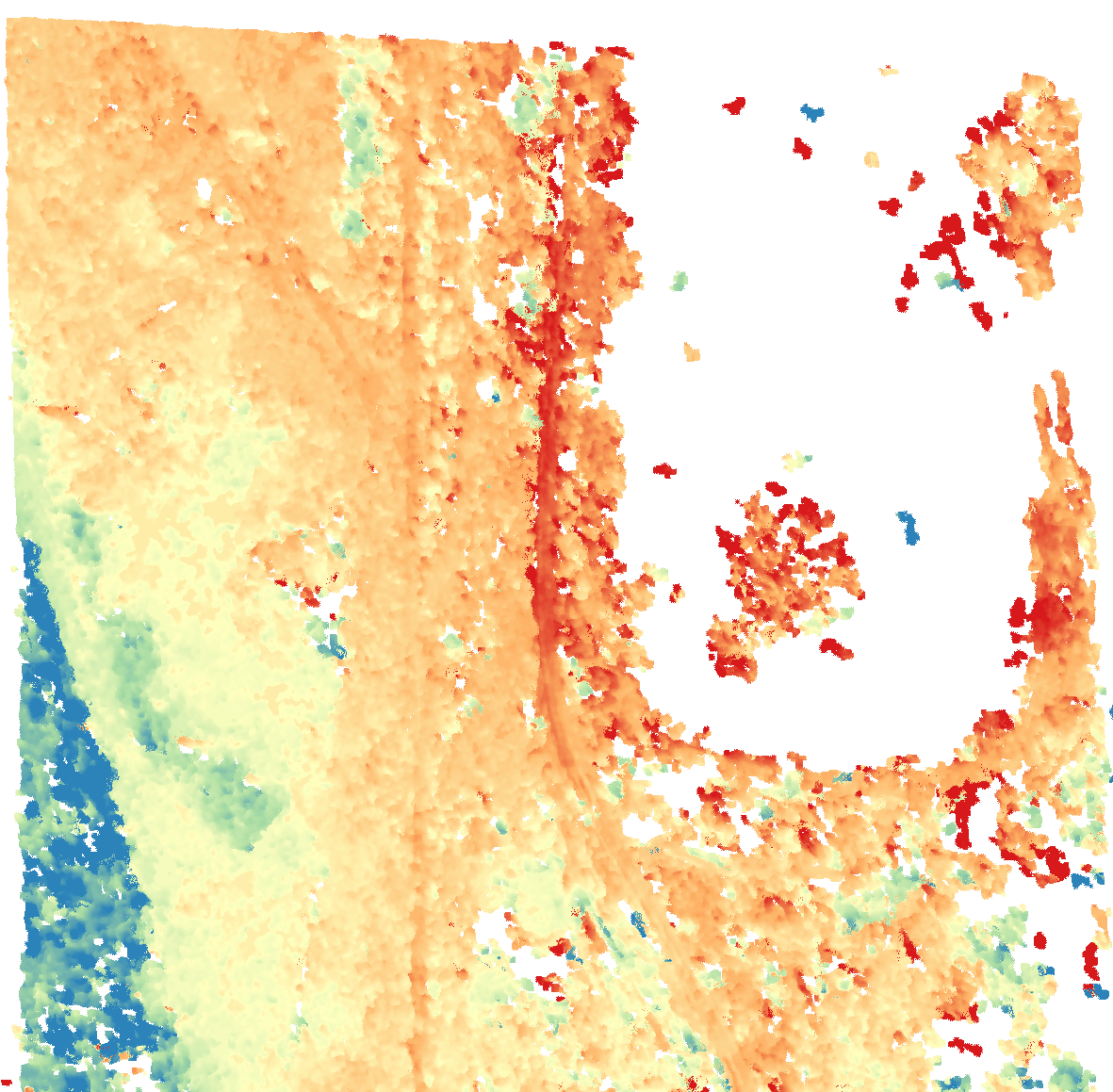}
    \caption{GIM-DKM}
    \label{fig:Comparison between DSM using relative orientation from matchers-h}
  \end{subfigure}
  \caption{Comparison between DSM using relative orientation from feature matching methods.}
  \label{fig:Comparison between DSM using relative orientation from matchers}
\end{figure}

\begin{figure}[hbt!]
  \centering
  \includegraphics[height=4.3cm]{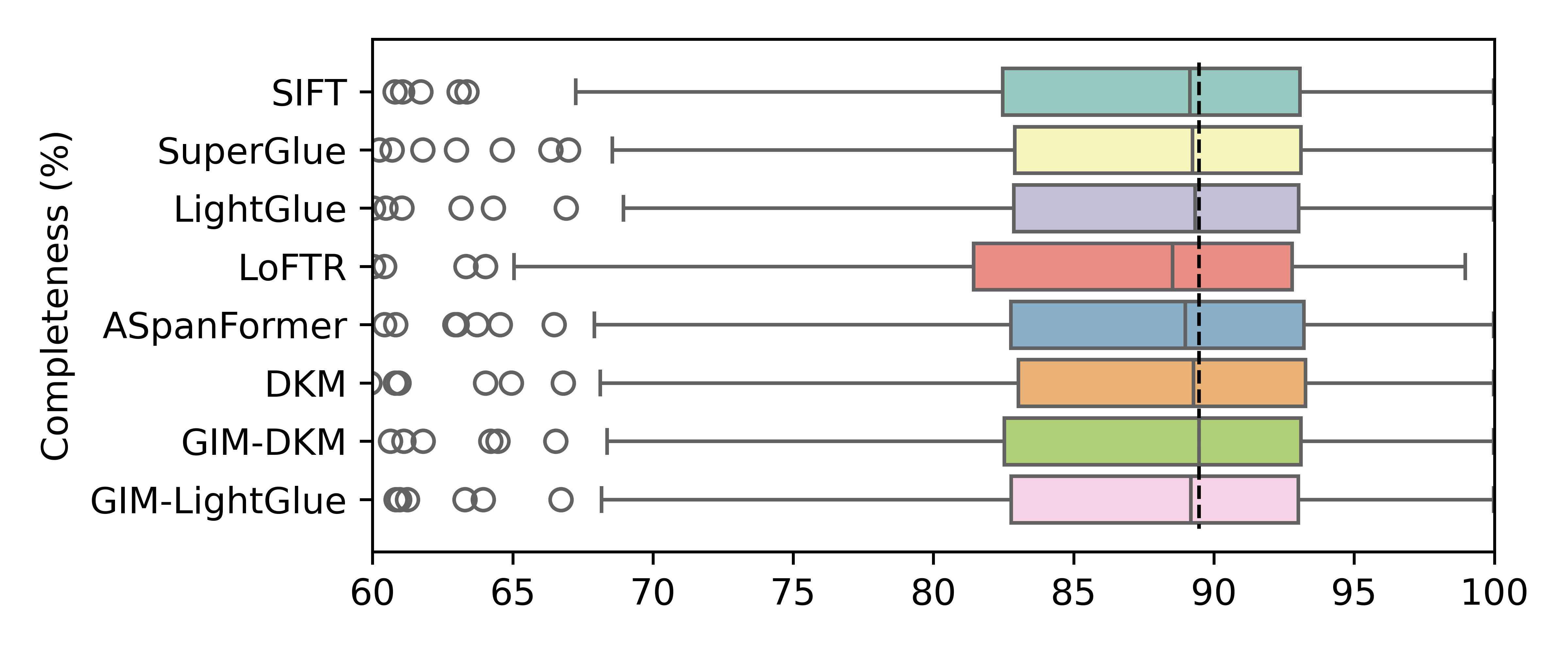}
  \caption{Comparison DSM \emph{completeness} of handcrafted and learning-based methods on all pairs where both methods successfully generated DSM in box plot. The vertical dash line indicates the best value.}
  \label{fig:Comparison DSM completeness}
\end{figure}

\begin{figure}[hbt!]
  \centering
  \includegraphics[height=4.3cm]{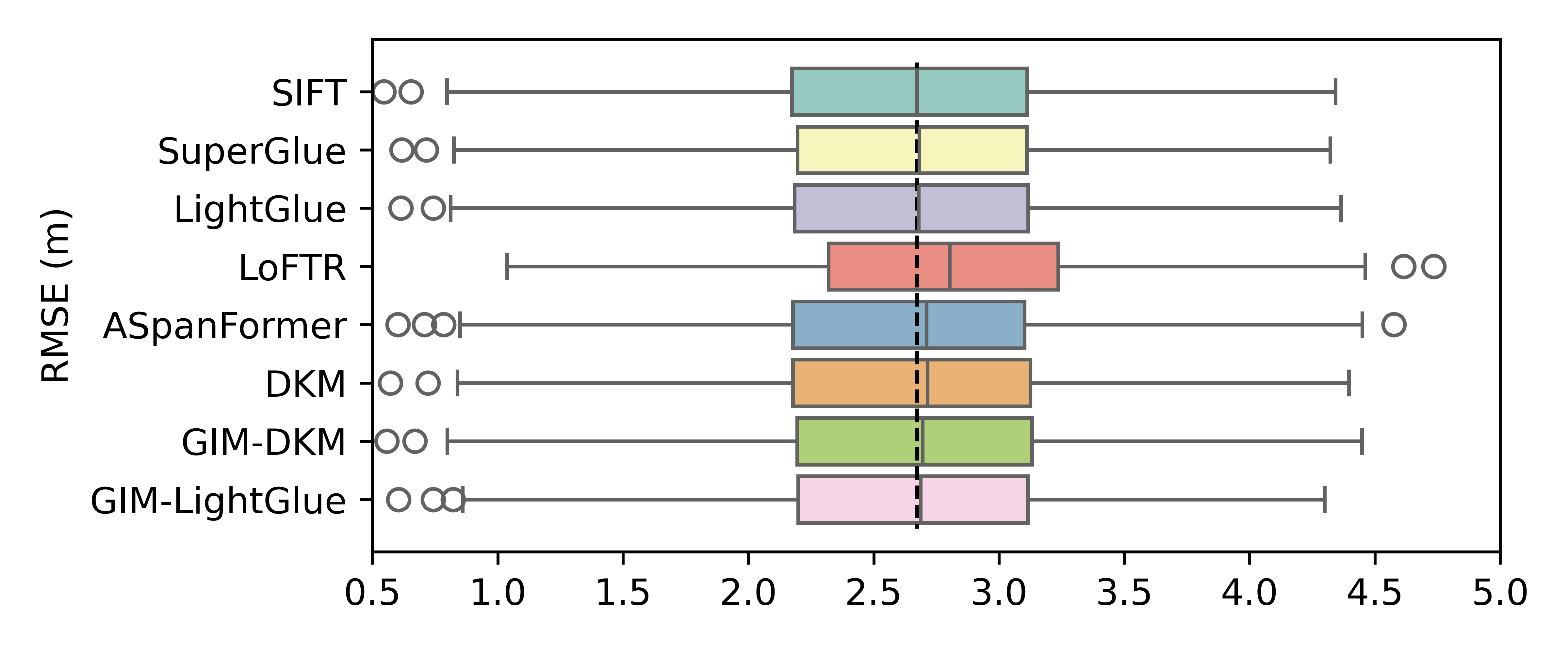}
  \caption{Comparison DSM quality of handcrafted and learning-based methods on all pairs where both methods successfully generated DSM in box plot. The vertical dash line indicates the best value.}
  \label{fig:Comparison DSM quality}
\end{figure}

\cref{fig:Comparison between DSM using relative orientation from matchers} compares DSMs produced with adjusted RPCs using handcrafted and learning-based matching methods. The completeness and accuracy of DSM are plotted in \cref{fig:Comparison DSM completeness}. In terms of completeness, DKM demonstrate the best completeness, meanwhile, LoFTR and ASpanFormer falls their rank. The rest including SIFT and other learning-based method shows similar performance in DSM completeness. \cref{fig:Comparison DSM quality} shows the final accuracy of the DSM reconstructed based on relative orientation computed with different methods comparing with ground truth. The performance of SIFT, SuperGlue, and LightGlue present the best DSM accuracy, and outperforms the rest of learning-based methods.

\subsection{Analysis of the Effectiveness of LSM for Point Localization Refinement}
\label{sec:Experiments4.4}

Least Squares Matching (LSM) \cite{bellavia_progressive_2024, bethmann_least-squares_2010, gruen_adaptive_1985} is a technique for patch-based point matching. It is often used to refine the positions of matched points to achieve sub-pixel accuracy for geometric processing, \ie, relative orientation or bundle adjustment. Considering that feature extraction may be performed on a low-resolution layer of the pyramid (such as SIFT), in our experiment, we explore the effectiveness of using LSM to enhance the accuracy of the matches by adjusting the point locations. We assess the relative change in evaluation metrics (refer to \cref{sec:Methodology3.4}) with and without LSM using \cref{eq:relative_change}.

\begin{equation}
  \text{Relative change} = \frac{m_{LSM} - m_{plain}}{m_{plain}} \times 100,
  \label{eq:relative_change}
\end{equation}
where $m$ is one of the previously defined metrics.

The relative changes (with and without applying the LSM) considers geometric processing statistics including inlier ratio, epipolar error, DSM completeness, and DSM accuracy across all pairs. The relative differences (by applying the LSM) are shown in \cref{tab:percentage_changes}. It can be seen that statistics can be improved notably when being refined by LSM, particularly regarding the quality of DSM, all methods might be benefit from LSM refinement.

\begin{table}[hbt!]
  \caption{The percentage changes in metrics due to applying LSM. For Inlier Ratio and DSM Completeness, larger values indicate better performance; whereas for epipolar error and DSM RMSE, smaller values are better.}
  \label{tab:percentage_changes}
  \centering
  \begin{tabular}{@{}l@{\hskip 0.2in}c@{\hskip 0.2in}c@{\hskip 0.2in}c@{\hskip 0.2in}c@{}}
    \toprule
    \bf Method & \makecell{\bf Inlier Ratio \\ (\%) $\uparrow$} & \makecell{\bf Epipolar Error \\ (\%) $\downarrow$} & \makecell{\bf Completeness \\ (\%) $\uparrow$} & \makecell{\bf RMSE \\ (\%) $\downarrow$} \\
    \midrule
    SIFT & 0.76 & -0.04 & 0.05 & -0.20 \\
    SuperGlue & \bf 4.37 & 0.10 & 0.07 & -0.47 \\
    LightGlue & 3.19 & 0.10 & 0.11 & \bf -0.72 \\
    LoFTR & -3.46 & -0.03 & 0.05 & -0.45 \\
    ASpanFormer & -4.55 & -0.06 & 0.10 & -0.53 \\
    DKM & -5.65 & \bf -0.17 & \bf 0.12 & -0.47 \\
    GIM-DKM & -5.50 & 0.01 & 0.10 & -0.61 \\
    GIM-LightGlue & 2.93 & 0.10 & 0.12 & -0.58 \\
    \bottomrule
  \end{tabular}
\end{table}

\section{Conclusions}
\label{sec:Conclusions}
This work evaluated the effectiveness of handcrafted and learning-based features for multi-date satellite stereo images. The evaluation focuses on geometric processing problems with off-track satellite stereo pairs. Using a large set of multi-date satellite images, we assessed the quality of matched points by evaluating the resulting accuracy of relative orientation and, subsequentially, the generated DSM. Our findings revealed that learning-based methods are generally superior in robustness of finding matchings than the handcrafted method. This was especially true in cases where the differences in sunlight and seasonal changes posed a challenge. However, for those cases where a handcrafted method is still able to find correspondences, their inliers are accurate in terms of photogrammetric processing. Considering the computational cost and scale-up capability, handcrafted matchers are still competitive in this age of deep learning. As learning-based results are promising, our future works aim to investigate the performance of other learning-based local features and matchers to support the extraction of geometric information from satellite off-track stereo pairs.


\section*{Acknowledgements}
The authors are supported in part by the Office of Naval Research [grant numbers N000142012141 \& N000142312670] and Intelligence Advanced Research Projects Activity (IARPA) via Department of Interior/ Interior Business Center (DOI/IBC) contract number 140D0423C0075.

%
%
\bibliographystyle{splncs04}
\bibliography{main}
\end{document}